\documentclass{article}

\usepackage{microtype}
\usepackage{graphicx}
\usepackage{subfigure}
\usepackage{booktabs} 

\usepackage{hyperref}

\usepackage{tikzit}

\usepackage[accepted]{icml2021}

\icmltitlerunning{Batch Value-function Approximation with Only Realizability}

\usepackage{url, hyperref}
\usepackage{amsthm, amsmath, amssymb} 
\usepackage[shortlabels]{enumitem}
\usepackage{comment}
\usepackage{color}
\usepackage{multirow}
\usepackage{xspace}
\usepackage{rotating}
\usepackage{graphicx}
\usepackage{caption}
\usepackage{xcolor}
\usepackage[normalem]{ulem}
\usepackage{tikz}

\usepackage{nicefrac}
\usepackage{mathtools}
\mathtoolsset{showonlyrefs,showmanualtags}

\theoremstyle{plain}
\newtheorem{theorem}{\textbf{Theorem}}
\newtheorem{lemma}[theorem]{\textbf{Lemma}}

\newtheorem{proposition}[theorem]{Proposition}

\newtheorem{example}{Example}

\theoremstyle{definition}
\newtheorem{definition}{Definition}
\newtheorem{assumption}{Assumption}

\renewcommand{\cite}{\citep}

\newcommand{\para}[1]{\noindent \textbf{#1}~}

\newcommand{\EE}{\mathbb{E}}

\newcommand{\VV}{\mathbb{V}}

\DeclareMathOperator*{\argmin}{arg\,min}
\DeclareMathOperator*{\argmax}{arg\,max}

\newcommand{\Fcal}{\mathcal{F}}
\newcommand{\Qcal}{\mathcal{Q}}

\newcommand{\Gcal}{\mathcal{G}}
\newcommand{\Hcal}{\mathcal{H}}
\newcommand{\Ncal}{\mathcal{N}}

\newcommand{\Scal}{\mathcal{S}}

\newcommand{\Acal}{\mathcal{A}}
\newcommand{\Tcal}{\mathcal{T}}

\newcommand{\Ecal}{\mathcal{E}}

\newcommand{\Rmax}{R_{\max}}

\newcommand{\Vmax}{V_{\max}}
\newcommand{\RR}{\mathbb{R}}
\newcommand{\Xcal}{\mathcal{X}}
\newcommand{\Ycal}{\mathcal{Y}}
\newcommand{\emp}[1]{\widehat{#1}}

\newcommand{\epsF}{\epsilon_{\Fcal}}
\newcommand{\epsphi}{\epsilon_\phi}
\newcommand{\epsphip}{\epsilon_{\phi'}}
\newcommand{\epsdct}{\epsilon_{\text{dct}}}
\newcommand{\epst}{\tilde{\epsilon}}
\newcommand{\epsd}{\epsilon_1}

\newcommand{\Fphi}{\Gcal_\phi}
\newcommand{\TG}{\Tcal_{\phi}^{\mu}}
\newcommand{\eTG}{\emp{\Tcal}_{\phi}^{\mu}}
\newcommand{\eTGp}{\emp{\Tcal}_{\phi'}^{\mu}}

\newcommand{\fh}{\hat{f}}

\newcommand{\CS}{C_{\Scal}}
\newcommand{\CA}{C_{\Acal}}
\newcommand{\ts}{\tilde{s}}
\newcommand{\ta}{\tilde{a}}

\newcommand{\Eff}{\mathcal{E}(f;f')}
\newcommand{\ff}{\hat{f}}

\newcommand{\fo}{f_0}

\begin{document}

\twocolumn[
\icmltitle{Batch Value-function Approximation with Only Realizability}

\icmlsetsymbol{equal}{*}

\begin{icmlauthorlist}
	\icmlauthor{Tengyang Xie}{uiuc}
	\icmlauthor{Nan Jiang}{uiuc}
\end{icmlauthorlist}

\icmlaffiliation{uiuc}{Department of Computer Science, University of Illinois at Urbana-Champaign, Illinois, USA}

\icmlcorrespondingauthor{Nan Jiang}{nanjiang@illinois.edu}

\icmlkeywords{batch reinforcement learning, realizability}

\vskip 0.3in
]

\printAffiliationsAndNotice{}  

\newcount\Comments  
\Comments=0 
\definecolor{darkgreen}{rgb}{0,0.5,0}
\definecolor{darkred}{rgb}{0.7,0,0}
\definecolor{teal}{rgb}{0.3,0.8,0.8}
\newcommand{\kibitz}[2]{\ifnum\Comments=1\textcolor{#1}{#2}\fi}
\newcommand{\nan}[1]{\kibitz{red}{[NJ: #1]}}
\newcommand{\TX}[1]{\kibitz{cyan}{[TX: #1]}}

\begin{abstract}
We make progress in a long-standing problem of batch reinforcement learning (RL): learning $Q^\star$ from an exploratory and polynomial-sized dataset, using a \emph{realizable} and otherwise \emph{arbitrary} function class. In fact, all existing algorithms demand function-approximation assumptions stronger than realizability, and the mounting negative evidence has led to a conjecture that sample-efficient learning is impossible in this setting  \citep{chen2019information}. Our algorithm, BVFT, breaks the hardness conjecture  (albeit under a stronger notion of exploratory data)  via a tournament procedure that reduces the learning problem to pairwise comparison, and solves the latter with the help of a state-action-space partition constructed from the compared functions. We also discuss how BVFT can be applied to model selection among other extensions and open problems.  
\end{abstract}

\section{Introduction} \label{sec:intro}
What is the minimal function-approximation assumption that enables polynomial sample complexity, when we try to learn $Q^\star$ from an exploratory batch dataset? Existing algorithms and analyses---those that have largely laid the theoretical foundation of modern reinforcement learning---have always demanded assumptions that are substantially stronger than the most basic one: \emph{realizability}, i.e., that $Q^\star$ (approximately) lies in the function class. These strong assumptions 
have recently compelled \citet{chen2019information} to conjecture an information-theoretic barrier, that polynomial learning is impossible in batch RL, even with exploratory data and realizable function approximation.

In this paper, we break this barrier by an algorithm called \textbf{B}atch \textbf{V}alue-\textbf{F}unction \textbf{T}ournament (BVFT). 
Via a tournament procedure, BVFT reduces the learning problem to that of identifying $Q^\star$ from a pair of candidate functions. In this subproblem, we create a piecewise constant function class of statistical complexity $O(1/\epsilon^2)$ that can express both candidate functions up to small discretization errors, and use the projected Bellman operator associated with the class to identify $Q^\star$. We present the algorithm in Section~\ref{sec:outline} and prove its sample complexity in Sections~\ref{sec:subproblem} and \ref{sec:tour}. 
A limitation of our approach is the use of a relatively stringent version of \emph{concentrability coefficient} from \citet{munos2003error} to measure the exploratoriness of the dataset (see Assumption~\ref{asm:con}). Section~\ref{sec:fail} investigates the difficulties in relaxing the assumption,  
and Appendix~\ref{app:violate} discusses how to mitigate the pathological behavior of the algorithm when the assumption does not hold. 

As another limitation, BVFT enumerates over the function class and is computationally inefficient for training. 
That said, the algorithm is efficient when the function class has a polynomial cardinality, making it applicable to another problem in batch RL: model selection \citep{farahmand2011model}.\footnote{We use the phrase ``model selection'' as in the context of e.g., cross validation, and the word ``model'' does not refer to MDP dynamics; rather they refer to value functions for our purposes.} 
In Section~\ref{sec:model_selection}, we review the literature on this important problem and discuss how BVFT has significantly advanced the state of the art on the theoretical front. 

\section{Related Work} \label{sec:related}
\para{Stronger Function-Approximation Assumptions in Existing Theory} The theory of batch RL has struggled for a long time to provide sample-efficiency guarantees when realizability is the only assumption imposed on the function class. An intuitive reason is that learning $Q^\star$ is roughly equivalent to minimizing the Bellman error, but the latter cannot be estimated from data \citep[Chapter 11.6]{jiang2019value, sutton2018reinforcement}, leading to the infamous ``double sampling'' difficulty  \citep{baird1995residual, antos2008learning}. Stronger/additional assumptions have been proposed to circumvent the issue, including low \emph{inherent Bellman errors} \citep{munos2008finite, antos2008learning}, \emph{averager} classes \citep{gordon1995stable}, and additional function approximation of importance weights \citep{xie2020q}. 

\para{State Abstractions} State abstractions are the simplest form of function approximation. (They are also special cases of the aforementioned averagers.) In fact, certainty equivalence with a state abstraction that can express $Q^\star$, known as \emph{$Q^\star$-irrelevant abstractions}, is known to be consistent, i.e., $Q^\star$ will be correctly learned if each abstract state-action pair receives infinite amount of data \citep[Theorem 4]{littman1996generalized, li2006towards}.  

While this observation is an important inspiration for our algorithm, making it useful for an arbitrary and unstructured function class is highly nontrivial and is one of the main algorithmic contributions of this paper. Furthermore, our finite-sample analysis significantly deviates from the ``tabular''-style  proofs in the abstraction literature \citep{paduraru2008model, nan_abstraction_notes}, where $\ell_\infty$ concentration bounds are established assuming that each abstract state receives sufficient data (c.f.~Footnote~\ref{ft:abs}). In our analysis, the structure of the abstraction is arbitrary, and it is much more convenient to treat them as piecewise constant classes over the original state space and use tools from  statistical learning theory to establish concentration results under weighted $\ell_2$ norm; see Section~\ref{sec:concentration} for details.

\para{Tournament Algorithms}
Our algorithm design also draws inspirations from existing tournament algorithms. Closest related is Scheff\'e tournament for density estimation \citep{devroye2012combinatorial}, which minimizes the total-variation (TV) distance from the true density among the candidate models, and has been applied to RL by \citet{sun2019model}. Interestingly, the main challenge in TV-distance minimization is very similar to ours at a high level, that TV-distance itself of a \emph{single} model cannot be estimated from data when the support of the distribution has a large or infinite cardinality. 
Similar to Scheff\'e tournament, our algorithm compares \emph{pairs} of candidate value functions, 
which is key to overcoming the fundamental unlearnability of Bellman errors.

Tournament algorithms are also found in RL  when the goal is to select the best state abstraction from a candidate set \citep{hallak2013model, jiang2015abstraction}. These works will be discussed in Section~\ref{sec:model_selection} in the context of model selection. 

\para{Lower Bounds} \citet{wang2020statistical, amortila2020variant, zanette2020exponential, chen2021infinite} have recently proved hardness results under $Q^\star$ realizability in batch RL. These results do not contradict ours because they deploy a weaker data assumption; see Appendix~\ref{app:lower} for discussions. Rather, their negative and our positive results are complementary and together provide a fine-grained characterization of the landscape of batch RL. 

\section{Preliminaries}
\subsection{Markov Decision Processes}
Consider an infinite-horizon discounted Markov Decision Process $(\Scal, \Acal, P, R, \gamma, d_0)$, where $\Scal$ is the finite state space that can be arbitrarily large, $\Acal$ is the finite action space, $P: \Scal\times\Acal\to\Delta(\Scal)$ is the transition function, $R: \Scal\times\Acal\to[0, \Rmax]$ is the reward function, $\gamma \in [0, 1)$ is the discount factor, and $d_0 \in \Delta(\Scal)$ is the initial state distribution. 

A (deterministic and stationary) policy $\pi: \Scal\to\Acal$ induces a distribution of the infinite trajectory $s_0, a_0, r_0, s_1, a_1, r_1, \ldots, $ as $s_0 \sim d_0, a_0 =\pi(s_0), r_0 = R(s_0, a_0), s_1 \sim P(s_0, a_0), \ldots$. We use $\EE[\cdot |\pi]$ to denote taking expectation w.r.t.~such a distribution.  The expected discounted return of a policy is defined as $J(\pi) := \EE[\sum_{t=0}^\infty \gamma^t r_t | \pi]$, and our goal is to optimize $J(\pi)$. Note that the random variable $\sum_{t=0}^\infty \gamma^t r_t$ is always bounded in the range $[0, \Vmax]$ where $\Vmax = \Rmax/(1-\gamma)$.

In the discounted setting, there is a policy $\pi^\star: \Scal\to\Acal$ that simultaneously optimizes the expected return for all starting states. This policy can be obtained as the greedy policy of the $Q^\star$ function, i.e., $\pi^\star(s) = \pi_{Q^\star}(s) := \argmax_{a\in\Acal} Q^\star(s,a)$, where we use $\pi_{(\cdot)}$ to denote a policy that greedily chooses actions according to a real-valued function over $\Scal\times\Acal$. The optimal Q-value function, $Q^\star$, can be uniquely defined through the Bellman optimality equations: $Q^\star = \Tcal Q^\star$, where $\Tcal: \RR^{\Scal\times\Acal} \to \RR^{\Scal\times\Acal}$ is the optimality operator,  defined as $(\Tcal f)(s,a) := R(s,a) + \gamma \EE_{s'\sim P(s,a)}[V_{f}(s',a')]$, where $V_f(s,a):= \max_{a} f(s,a)$. 

\subsection{Batch Data}  \label{sec:data}
We assume that the learner has access to a batch dataset $D$ consisting of i.i.d.~$(s,a,r,s')$ tuples, where $(s,a)\sim \mu, r=R(s,a), s'\sim P(s,a)$. Such an i.i.d.~assumption is standard for finite-sample analyses in the ADP literature \citep{munos2008finite, farahmand2010error, chen2019information}, and can often be relaxed at the cost of significant technical burdens and complications \citep[see e.g.,][]{antos2008learning}. 
We will also use $\mu(s)$ and $\mu(a|s)$ to denote the marginal of $s$ and the conditional of $a$ given $s$. To learn a near-optimal policy in batch RL, an exploratory dataset is necessary, and we measure the degree of exploration as follows:
\begin{assumption} \label{asm:con}
We assume that $\mu(s,a) > 0~ \forall s,a$. 
We further assume that \\
(1) There exists constant $1 \le \CA< \infty$ such that for any $s\in\Scal, a\in\Acal$, $\mu(a|s) \ge 1/\CA$. \\
(2) There exists constant $1 \le \CS < \infty$ such that  for any $s\in\Scal, a\in\Acal, s'\in\Scal$, $P(s'|s,a)  / \mu(s') \le \CS$. Also $d_0(s) / \mu(s) \le \CS$. \\
It will be  convenient to define $C = \CS \CA$. 
\end{assumption}
The first statement is very standard, asserting that the data distribution put enough probabilities on all actions. For example, with a small number of actions, a uniformly random policy ensures that $\CA = |\Acal|$ satisfies this assumption.

The second statement measures the exploratoriness of $\mu$'s state marginal by $\CS$, and two comments are in order. First, this is a form of \emph{concentrability} assumption, which not only enforces data to be exploratory, but also implicitly imposes restrictions on the MDP's dynamics (see the reference to $P$ in Assumption~\ref{asm:con}). While the latter may be undesirable, \citet[Theorem 4]{chen2019information} shows that such a restriction is \emph{unavoidable} when learning with a general function class. Second, the version of concentrability coefficient we use was introduced by \citet[Eq.(6)]{munos2003error}, and is more stringent than its more popular variants  \citep[e.g.,][]{munos2007performance, farahmand2010error}. That said, (1) hardness results exist under a weaker form of the assumption (see Appendix~\ref{app:lower}), and (2) whenever the transition dynamics admit low-rank stochastic factorization, there always exist data distributions that yield small $\CS$ despite that $|\Scal|$ can be arbitrarily large; see Appendix~\ref{app:CS}, where we also discuss how Assumption~\ref{asm:con} compares to no inherent Bellman errors in the context of low-rank MDPs.  We investigate why it is difficult to work with more relaxed assumptions in Section~\ref{sec:fail}, and discuss how to mitigate the negative consequences when the assumption is violated in Appendix~\ref{app:violate}.

A direct consequence of Assumption~\ref{asm:con}, which we will use later to control error propagation and distribution shift, is the following proposition.
\begin{proposition}\label{prop:explore}
Let $\nu$ be a distribution over $\Scal\times\Acal$ and $\pi$ be a policy. Let $\nu' = P(\nu) \times \pi$ denote the distribution specified by the generative process $(s',a') \sim \nu' \Leftrightarrow (s,a) \sim \nu, s'\sim P(\cdot|s,a), a' = \pi(s')$. Under Assumption~\ref{asm:con}, we have $\|\nu'/\mu\|_\infty := \max_{s, a} \nu'(s,a)/\mu(s,a) \le C$.  Also note that $\|(d_0 \times \pi) / \mu\|_\infty \le C$.
\end{proposition}

\para{Additional Notations} For any real-valued function of $(s,a,r,s')$, we use $\EE_\mu[\cdot]$ as a shorthand for taking expectation of the function when $(s,a) \sim \mu, r= R(s,a), s'\sim P(s,a)$. Also for any $f:\Scal\times\Acal\to\RR$, define $\|f\|_{2, \mu}^2 := \EE_{\mu}[f^2]$; $\|f\|_{2, \mu}$ is a weighted $\ell_2$ norm and satisfies the triangular inequality. We also use $\|f\|_{2, D}^2$ to denote the empirical approximation of $\|f\|_{2, \mu}^2$ based on the dataset $D$. 

\subsection{Value-function Approximation}
Since the state space $\Scal$ can be prohibitively large, function approximation is necessary for scaling RL to large and complex problems. In the value-function approximation setting, we are given a function class $\Fcal \subset (\Scal\times\Acal \to [0, \Vmax])$ to model $Q^\star$. Unlike prior works that measure the approximation error of $\Fcal$ using \emph{inherent Bellman errors} \citep{munos2007performance, antos2008learning}---which amounts to assuming that $\Fcal$ is (approximately) \emph{closed} under $\Tcal$---we will measure the error using Definition~\ref{def:approx}, where $0$ error \emph{only} implies realizability, $Q^\star\in\Fcal$. In fact, given that the assumptions required by all existing algorithms are substantially stronger than realizability, \citet[Conjecture 8]{chen2019information} conjecture that polynomial sample complexity is unattainable in  batch RL when we only impose realizability on $\Fcal$, which is why our result may be surprising.

\begin{definition} \label{def:approx}
Let $\epsF := \inf_{f \in \Fcal} \|f - Q^\star\|_{\infty}$. \footnote{It is possible to define $\epsF$ under weighted $\ell_2$ norm, though making this change in our current proof yields a worse (albeit polynomial) sample complexity; see Appendix~\ref{app:l2approx} for more details.} Let $f^\star$ denote the $f$ that attains the infimum. 
\end{definition}
We assume $\Fcal$ is finite but exponentially large (as in \citet{chen2019information}), i.e., we can only afford $\text{poly}\log|\Fcal|$ in the sample complexity. For continuous function classes that admit a finite $\ell_\infty$ covering number \citep{agarwal2011covering}, our approach and analysis immediately extend by replacing $\Fcal$ with its $\epsilon$-net at the cost of slightly increasing $\epsF$.

\subsection{Polynomial Learning}
Our goal is to devise a statistically efficient algorithm with the following kind of guarantee: with high probability we can learn an $\epsilon$-optimal policy $\hat{\pi}$, that is, $J(\hat{\pi}) \ge J(\pi^\star) - \epsilon \cdot \Vmax$, when $\Fcal$ is realizable and the dataset $D$ is only polynomially large. The polynomial may depend on the effective horizon $1/(1-\gamma)$, the statistical complexity of the function class $\log|\Fcal|$, the concentrability coefficient $C$, (the inverse of) the suboptimality gap $\epsilon$, and $1/\delta$ where $\delta$ is the failure probability. 
Our results can also accommodate the more general setting when $\Fcal$ is not exactly realizable, in which case the suboptimality of $\hat{\pi}$ is allowed to contain an additional term proportional to the approximation error $\epsF$ up to a polynomial multiplicative factor.

\begin{algorithm}[t]
	\caption{\textbf{B}atch \textbf{V}alue-\textbf{F}unction \textbf{T}ournament (BVFT)}
	\label{alg:bvft}
	\begin{algorithmic}[1]
		\STATE \textbf{Input:} Dataset $D$, function class $\Fcal$, discretization parameter $\epsdct \in (0, \Vmax)$. 
		\FOR{$f\in\Fcal$} 
		\STATE $\bar{f} \gets $ discretize the output of $f$ with resolution $\epsdct$. (see Footnote~\ref{ft:discrete}). 
		\ENDFOR 
		\FOR{$f\in\Fcal$} \label{lin:f}
		\FOR{$f'\in\Fcal$}  \label{lin:f1}
		\STATE Define $\phi$ s.t.~$\phi(s,a) = \phi(s',a')$ iff $\bar f(s,a)=\bar f(s',a')$ and $\bar f'(s,a) = \bar f'(s',a')$.  \label{lin:phi}
		\STATE $\Eff \gets \|f - \eTG f \|_{2, D}$  (see Eq.\eqref{eq:eTG} for def of $\eTG$; the dependence on $f'$ is only through $\phi$). \label{lin:Eff}
		\ENDFOR
		\ENDFOR
		\STATE $\ff \gets \argmin_{f\in\Fcal} \max_{f'\in\Fcal}\Eff$.
		\STATE \textbf{Output:} $\hat{\pi} = \pi_{\ff}$.
	\end{algorithmic}
\end{algorithm}

\section{Algorithm and the Guarantee}  \label{sec:outline}
In this section, we introduce and provide intuitions for our algorithm, and state its sample complexity guarantee which will be proved in the subsequent sections.

The design of our algorithm is based on an important observation inspired by the state-abstraction literature (see Section~\ref{sec:related}): when the function class $\Fcal$ is piecewise constant and realizable, batch learning with exploratory data is consistent using e.g., Fitted Q-Iteration. This is because piecewise constant classes are very stable, and their associated projected Bellman operators are always $\gamma$-contractions under $\ell_\infty$, implying that $Q^\star$ is the only fixed point of such operators when statistical errors are ignored;\footnote{$Q^\star$ is always \emph{a} fixed point of the projected Bellman update operators associated with any realizable function class, but there is no uniqueness guarantee in general.} we will actually establish these properties in Section~\ref{sec:exact}. 

While realizability is the only expressivity condition assumed, being piecewise constant is a major structural assumption, and is too restrictive to accommodate practical function-approximation schemes such as linear predictors or neural networks, let alone the completely unstructured set of functions one would encounter in model selection (Section~\ref{sec:model_selection}). How can we make use of this observation? 

An immediate idea is \emph{improper learning}, i.e., augmenting  $\Fcal$---which is not piecewise constant in general and may have an arbitrary structure---to its smallest superset that \emph{is} piecewise constant, which automatically inherits realizability from $\Fcal$. To do so, we may first discretize the output of each function $f \in \Fcal$ up to a small discretization error $\epsdct$,\footnote{When $\Vmax/\epsdct$ is an odd integer, discretization onto a regular grid $\{\epsdct, 3\epsdct, \ldots, \Vmax-\epsdct\}$ guarantees at most $\epsdct$ approximation error, and the cardinality of the set is $\Vmax/2\epsdct$. For arbitrary $\epsdct \in (0, \Vmax)$, a similar discretization yields  a cardinality of $\lceil\Vmax/2\epsdct \rceil$, and we upper-bound it by $\Vmax/\epsdct$ throughout the analysis for convenience. \label{ft:discrete}} and partition $\Scal\times\Acal$ by grouping state-action pairs together \emph{only when the output $f\in\Fcal$ (after discretization) is constant across them}. The problem is that, the resulting function class is way too large compared to $\Fcal$; its statistical complexity---measured by the number of groups---can be as large as $(\Vmax/\epsdct)^{|\Fcal|}$, \emph{doubly exponential} in $\text{poly}\log|\Fcal|$ which is what we can afford!

To turn this idea into a polynomial algorithm, we note that the statistical complexity of the superset is affordable when $|\Fcal|$ is constant, say, $|\Fcal|=2$. This provides us with a procedure that identifies $Q^\star$ out of two candidate functions. To handle an exponentially large $\Fcal$, we simply perform pairwise comparisons between all pairs of $f,f'\in\Fcal$, and output the function that has survived all pairwise comparisons involving it. 
Careful readers may wonder what happens when $Q^\star\notin\{f, f'\}$, as realizability is obviously violated. As we will show in Section~\ref{sec:tour}, the outcomes of these ``bad'' comparisons simply do not matter: $Q^\star$ is never involved in such comparisons, and any other function $f$ will always be checked against $f'= Q^\star$, which is enough to expose the deficiency of a bad $f$. 

The above reasoning ignores approximation and estimation errors, which we handle in the actual algorithm and its analysis; see Algorithm~\ref{alg:bvft}. Below we state its sample complexity guarantee, which is the main theorem of this paper. 
\begin{theorem}\label{thm:main}
Under Assumption~\ref{asm:con}, with probability at least $1-\delta$, BVFT (Algorithm~\ref{alg:bvft}) with $\epsdct = \frac{(1-\gamma)^2 \epsilon \Vmax}{16\sqrt{C}}$ returns a policy $\hat{\pi}$ that satisfies
\begin{align} \label{eq:opt}
J(\pi^\star) - J(\hat\pi) \le \frac{(4+8\sqrt{C})\epsF}{(1-\gamma)^2} + \epsilon\cdot \Vmax,
\end{align}
with  a sample complexity of \footnote{$\tilde O(\cdot)$ suppresses poly-logarithmic dependencies.}
\begin{align}
|D| = \tilde{O}\left(\frac{C^2 \ln\frac{|\Fcal|}{\delta}}{\epsilon^4 (1-\gamma)^8}\right). 
\end{align}
\end{theorem}
The most outstanding characteristic of the sample complexity is the $1/\epsilon^4$ rate. In fact, the poor dependencies on $C$ and $1/(1-\gamma)$ are both due to $1/\epsilon^4$: when we rewrite the guarantee in terms of suboptimality gap as a function of $n=|D|$, we see an $O(\sqrt{C} n^{-1/4}/(1-\gamma)^2)$ estimation-error term, featuring the standard $\sqrt{C}$ penalty due to distribution shift and quadratic-in-horizon error propagation. 

The $1/\epsilon^4$ rate comes from two sources: $1/\epsilon^2$ of it is due to the worst-case statistical complexity of the piecewise constant classes created during pairwise comparisons. The other $1/\epsilon^2$ is the standard statistical rate. While standard, proving $O(1/\epsilon^2)$ concentration bounds in our analysis turns out to be technically challenging and requires some clever tricks. We refer mathematically inclined readers to Section~\ref{sec:concentration} for how we overcome those challenges. 


We prove Theorem~\ref{thm:main} in the next two sections. Section~\ref{sec:subproblem} establishes the essential properties of the pairwise comparison step in Line~8, where we view the problem at a somewhat abstract level to attain proof modularity. Section~\ref{sec:tour} uses the results in Section~\ref{sec:subproblem} to prove the final guarantee. 

\section{Value-function Validation using a Piecewise Constant Function Class}
\label{sec:subproblem}
In this section we analyze a subproblem that is crucial to our algorithm: given a piecewise constant class $\Fphi \subset (\Scal\times\Acal\to[0, \Vmax])$ (induced by $\phi$, a partition of $\Scal\times\Acal$)\footnote{We treat $\phi$ as  mapping $\Scal\times\Acal$ to an arbitrary finite codomain, and $g(s,a) = g(s',a')~\forall g\in\Fphi$ iff $\phi(s,a) = \phi(s',a')$.} with small realizability error $\epsphi := \epsilon_{\Fphi}$, 
we show that we can compute a statistic for any given function $\fo: \Scal\times\Acal\to[0, \Vmax]$, and the statistic will be a good surrogate for $\| \fo- Q^\star\|$ as long as Assumption~\ref{asm:con} holds and the sample size is polynomially large. 
We use $|\phi|$ to denote the number of equivalence classes induced by $\phi$. 

As Section~\ref{sec:outline} and Algorithm~\ref{alg:bvft} have already alluded to, later we will invoke this result when comparing two candidate value functions $f$ and $f'$ (with $\fo = f$), and define $\phi$ as the coarsest partition that can express both $f$ and $f'$; when $Q^\star \in \{f, f'\}$, $\epsphi$ will be small.  To maintain the modularity of the analysis, however, we will view $\phi$ as an arbitrary partition of $\Scal\times\Acal$ in this section. 

The statistic we compute is $\|\fo - \eTG \fo\|_{2, D}$ (c.f.~Line~8 of Algorithm~\ref{alg:bvft}), where $\eTG$ is defined as follows:
\begin{definition}
Define $\eTG$ as the sample-based projected Bellman update operator associated with $\Fphi$: for any $f: \Scal\times\Acal\to[0, \Vmax]$, ~~~$\eTG f:= $ 
\begin{align} \label{eq:eTG}
\argmin_{g\in\Fphi} \frac{1}{|D|}\sum_{(s,a,r,s') \in D} [(g(s,a) - r - \gamma V_f(s'))^2].
\end{align}
\end{definition}

\subsection[Warm up: |D| to infty and epsphi = 0]{Warm up: $|D|\to\infty$ and $\epsphi = 0$} \label{sec:exact}
To develop intuitions, we first consider  the special case of $|D|\to\infty$ and $\epsphi = 0$. In this scenario, we can show that $Q^\star$ is the unique fixed point of $\eTG$, which justifies using $\|\fo - \eTG \fo\|$ as a surrogate for $\|\fo - Q^\star\|$. The concepts and lemmas introduced here will also be  useful  for the later analysis of the general case. 

We start by defining $\TG$ as   $\eTG$ when $|D|\to\infty$.
\begin{definition} \label{def:TG}
Define $\TG$ as the projected Bellman update where the projection is onto $\Fphi$, weighted by $\mu$. That is, for any $f: \Scal\times\Acal\to [0, \Vmax]$,
\begin{align} \label{eq:TG}
\TG f := \argmin_{g\in\Gcal_\phi} \EE_{\mu}[(g(s,a) - r - \gamma V_f(s'))^2].
\end{align}
\end{definition}
Next, we show that it is possible to define an MDP $M_\phi$, such that $\TG$ coincides with the Bellman update of $M_\phi$. 
Readers familiar with state abstractions may find the definition unusual, as  the ``abstract MDP'' associated with $\phi$ is typically defined over the compressed (or abstract) state space instead of the original one \citep[e.g.,][]{ravindran2004approximate}. We define $M_\phi$ over $\Scal$ because (1) our $\phi$ is an arbitrary partition of $\Scal\times\Acal$, which does not necessarily induce a consistent notion of abstract states, and (2) even when it does, the MDPs defined over $\Scal$ are dual representations to and share many important properties with the classical notion of abstract MDPs \citep{nan_abstraction_notes}. 
\begin{definition}\label{def:Mphi}
Define $M_\phi = (\Scal, \Acal, P_\phi, R_\phi, \gamma, d_0)$, where  
\begin{align*}
& R_{\phi}(s,a) = \frac{\sum_{\ts, \ta: \phi(\ts, \ta) = \phi(s,a)} \mu(\ts, \ta) R(\ts, \ta)}{\sum_{\ts, \ta: \phi(\ts, \ta) = \phi(s,a)} \mu(\ts, \ta) }. \\
& P_{\phi}(s'|s,a) = \frac{\sum_{\ts, \ta: \phi(\ts, \ta) = \phi(s,a)} \mu(\ts, \ta) P(s'|\ts, \ta)}{\sum_{\ts, \ta: \phi(\ts, \ta) = \phi(s,a)} \mu(\ts, \ta) }.
\end{align*}
\end{definition}

\begin{lemma} \label{lem:proj_bellman}
$\TG$ is the Bellman update operator of $M_\phi$. 
\end{lemma}
Lemma~\ref{lem:proj_bellman} implies that $\TG$ is a $\gamma$-contraction under $\ell_\infty$ and has a unique fixed point, namely the optimal $Q$-function of $M_\phi$. It then suffices to show that $Q^\star$ is such a fixed point. 

\begin{proposition} \label{prop:epsphi0}
When $\epsphi=0$, $Q^\star$ is the unique fixed point of $\TG$.
\end{proposition}

\subsection{The General Case}
In the general case, we want to show that $\|\fo - \eTG \fo \|_{2, D}$ and $\|\fo - Q^\star\|$ control each other. 
The central result of this section is the following proposition:

\begin{proposition} \label{prop:phi}
Fixing any $\epsd, \epst$. Suppose
\begin{align}\label{eq:sample_size_phi}
|D| \ge \frac{32\Vmax^2 |\phi|\ln\frac{8\Vmax}{\epst \delta}}{\epst^2} + \frac{50 \Vmax^2 |\phi| \ln{\frac{80 \Vmax}{\epsd \delta}}}{\epsd^2}.
\end{align}
Then, with probability at least $1-\delta$, 
for any $\nu\in \Delta(\Scal\times\Acal)$ such that $\|\nu/\mu\|_{\infty} \le C$,  
\begin{align} \label{eq:nobad}
\textstyle \|\fo - Q^\star \|_{2,\nu} \le \frac{2\epsphi + \sqrt{C}(\|\fo - \eTG \fo \|_{2, D} + \epsd + \epst)}{1-\gamma}. 
\end{align}
At the same time, 
\begin{align} \label{eq:hasgood}
\|\fo - \eTG \fo\|_{2, D} \le (1+\gamma)\|\fo - Q^\star\|_\infty + 2\epsphi + \epst + \epsd. 
\end{align}
 \end{proposition}

Proving the proposition requires quite some preparations. We group the helper lemmas according to their nature in Sections~\ref{sec:err_prop} to \ref{sec:concentration}, and prove Proposition~\ref{prop:phi} in Section~\ref{sec:phiproof}.

\subsubsection{Error Propagation} \label{sec:err_prop}
The first two lemmas allow us to characterize error propagation in later proofs. That is, it will help answer the question: if we find $\|\fo - \TG \fo\|_{2, \mu}$ to be small (but nonzero), why does it imply that $\|\fo - Q^\star\|$ is small? 

While results of similar nature exist in the state-abstraction literature, they often bound $\|\fo - Q^\star\|$ with $\|\fo -\TG \fo\|_{\infty}$, where error propagation is easy to handle \citep{nan_abstraction_notes}. However, $\|\fo -\TG \fo\|_{\infty}$ can only be reliably estimated if each group of state-action pairs receives a sufficient portion of the data, which is not guaranteed in our setting due to the arbitrary nature of $\phi$ created in Line~7. This forces us to work with the $\mu$-weighted $\ell_2$-norm and  carefully characterize how error propagation shifts the distributions. 

In fact, it is precisely this analysis that demands the strong definition of concentrability coefficient $\CS$ in Assumption~\ref{asm:con}: as we will later show in the proof of Proposition~\ref{prop:phi} (Section~\ref{sec:phiproof}), the error propagates according to the dynamics of $M_\phi$ instead of that of $M$ (c.f.~the $P_\phi(\nu)$ term in Eq.\eqref{eq:error_prop}). Therefore, popular definitions of concentrability coefficient \citep[e.g.,][]{munos2007performance, antos2008learning, farahmand2010error, xie2020q}---which all consider state distributions induced in $M$---do not fit our analysis. 
Fortunately, the $\CS$ defined in Assumption~\ref{asm:con} 
has a very nice property, that it automatically carries over to $M_\phi$ no matter what $\phi$ is: 
\begin{lemma} \label{lem:Mphicon}
Any $C< \infty$ that satisfies Assumption~\ref{asm:con} for the true MDP $M$ also satisfies the same assumption in $M_\phi$. As a further consequence, Proposition~\ref{prop:explore} is also satisfied when $P$ is replaced by $P_\phi$.
\end{lemma}


\subsubsection[Error of Q* under T]{Error of $Q^\star$ under $\TG$} \label{sec:apprx}
The next lemma parallels Proposition~\ref{prop:epsphi0} in Section~\ref{sec:exact}, where we showed that $\|Q^\star - \TG Q^\star\|=0$ when $\epsphi = 0$. When $\epsphi$ is non-zero, we need a more robust version of this result showing that $\|Q^\star - \TG Q^\star\|$ is controlled by $\epsphi$. 
\begin{lemma} \label{lem:qapprox}
	$\|Q^\star - \TG Q^\star\|_{\infty} \le 2\epsphi$.
\end{lemma}

\subsubsection{Concentration Bounds} \label{sec:concentration}
We need two concentration events: that $\eTG \fo$ is close to $\TG \fo$, and that $\|\fo - \eTG \fo\|_{2, D}$ is close to $\|\fo - \eTG \fo\|_{2, \mu}$. We will split the failure probability $\delta$ evenly between these events. 

\paragraph{Concentration of $\eTG \fo$} We begin with the former, which requires a standard result for realizable least-square regression. The proof is deferred to Appendix~\ref{app:least_square}.
\begin{lemma}[Concentration Bound for Least-Square Regression]\label{lem:regression}
	Consider a real-valued regression problem with feature space $\Xcal$ and label space $\Ycal \subset[0, \Vmax]$. Let  $(x_i, y_i) \sim P_{X, Y}$ be $n$ i.i.d.~data points. Let $\Hcal \subset (\Xcal \to \Ycal)$ be a hypothesis class with $\ell_\infty$ covering number $N = \Ncal_\infty(\Hcal, \epsilon_0)$ and that realizes the  Bayes-optimal regressor, i.e.,  $h^\star = (x \mapsto \EE[Y|X=x]) \in \Hcal$. Let $\hat h = \argmin_{h\in\Hcal} \hat{\EE}[(h(X)-Y)^2]$ be the empirical risk minimizer (ERM), where $\hat{\EE}$ is the empirical expectation based on $\{(x_i, y_i)\}_{i=1}^n$. Then, with probability at least $1-\delta$, 
	$$
	\EE[(h^\star(X) - \hat h(X))^2] \le \frac{8\Vmax^2 \log\frac{N}{\delta}}{n} + 8\Vmax\epsilon_0.
	$$
\end{lemma}
We then use Lemma~\ref{lem:regression} to prove that $\|\eTG f - \TG f\|_{2, \mu}$ is small.

\begin{lemma} \label{lem:generr}
	Fixing $f: \Scal\times\Acal\to[0, \Vmax]$. W.p.~$\ge 1-\tfrac{\delta}{2}$, 
	$\|\eTG f - \TG f\|_{2, \mu} \le \epst$, as long as 
	\begin{align} \label{eq:subproblem_sample_size}
	|D| \ge \frac{16\Vmax^2 (2|\phi|\log(4\Vmax/\epst)+ \log(2/\delta))}{\epst^2}.
	\end{align}
\end{lemma}

\paragraph{Technical Challenge \& Proof Idea} Recall that $\eTG f$ is the ERM (in $\Fphi$) of the least-square regression problem $(s,a) \mapsto r+\gamma V_f(s')$, so $\|\eTG f - \TG f\|_{2, \mu}$ essentially measures the $\mu$-weighted $\ell_2$ distance between the ERM and the population risk minimizer. Proving this is straightforward when the regression problem is realizable, as  Lemma~\ref{lem:regression} would be directly applicable.\footnote{See e.g.,~Lemma 16 of \citet{chen2019information}, where the (approximate) realizability of any such regression problem is guaranteed by the assumption of low inherent Bellman error.} In our case, however, the regression problem is in general non-realizable (except for $f = Q^\star$) and can incur arbitrarily large approximation errors, as $\Fphi$ does not necessarily contain the Bayes-optimal regressor $\Tcal f$. 

The key proof idea is to leverage a special property of piecewise constant classes\footnote{An alternative (and much messier) approach is to prove scalar-valued concentration bounds for $\eTG f$ in each group of state-action pairs. Those groups with few data points will have high uncertainty, but they also contribute little to $\|\cdot\|_{2, \mu}$. Compared to this approach, our proof is much simpler. \label{ft:abs}} to reduce the analysis to the realizable case: regressing $(s,a) \mapsto r + \gamma V_g(s')$ over $\Fphi$ is equivalent to regressing $x \mapsto r + \gamma V_g(s')$ (with $x = \phi(s,a)$) over a  ``tabular'' function class, where the $s,a|x$ portion of the data generation process is treated as part of the inherent label noise. After switching to this alternative view, the tabular class over the codomain of $\phi$ is fully expressive and always realizable, which makes Lemma~\ref{lem:regression} applicable. See Appendix~\ref{app:generr} for the full proof of Lemma~\ref{lem:generr}.

\paragraph{Concentration of $\|\fo - g\|_{2,D}$}  The second concentration result we need is an upper bound on $|\|\fo - g\|_{2, D} - \|\fo - g\|_{2, \mu}|$ for all $g\in\Fphi$ simultaneously. We need to union bound over $g\in\Fphi$ because our statistic is $\|\fo-g\|_{2, D}$ with $g=\eTG \fo$, which is a data-dependent function. 

\begin{lemma} \label{lem:distance_gen_err}
W.p.~$\ge 1-\delta/2$,  $\forall g\in \Fphi$, $|\|\fo - g\|_{2, D} - \|\fo - g\|_{2, \mu}| \le \epsd$, as long as 
\begin{align} \label{eq:distance_sample_size}
|D| \ge \frac{50 \Vmax^2 |\phi| \ln{\frac{80 \Vmax}{\epsd \delta}}}{\epsd^2}.
\end{align}
\end{lemma}
\paragraph{Technical Challenge  \& Proof Idea} It is straightforward to bound $|\|\fo - g\|_{2, D}^2 - \|\fo - g\|_{2, \mu}^2|$ (note the squares), but a na\"ive conversion to a bound on the desired quantity (difference without squares) would result in $O(n^{-1/4})$ rate. 
To obtain $O(n^{-1/2})$ rate, we consider two situations separately, depending on whether  $\|\fo - g\|_{2, \mu}$ is below or above certain threshold: when it is below the threshold, we can use Bernstein's to exploit the low variance of $(\fo-g)^2$; when it is above the threshold, we obtain the bound by factoring the difference of squares. Combining these two cases with an  $O(\epsd)$ threshold yields a clean $O(n^{-1/2})$ result; see proof details in Appendix~\ref{app:distance_gen_err}. 

\subsubsection{Proof of Proposition~\ref{prop:phi}} \label{sec:phiproof}
We are now ready to prove Proposition~\ref{prop:phi}. Due to space limit we only provide a proof sketch in the main text.
\begin{proof}[Proof Sketch]
To prove Eq.\eqref{eq:nobad}, define $\pi_{f, f'}$ as the policy $s \mapsto \argmax_{a} \max\{f(s,a), f'(s,a)\}$. Consider any $\nu$ such that $\|\nu/\mu\|_\infty \le C$, we have $\|Q^\star - \fo\|_{2, \nu} \le$
\begin{align*}
 \|Q^\star - \TG Q^\star\|_{2, \nu} + \|\TG Q^\star - \TG \fo\|_{2, \nu} +  \|\fo- \TG \fo\|_{2, \nu}.
\end{align*}
The first term can be bounded via Lemma~\ref{lem:qapprox}. The second term is bounded by $\gamma \|Q^\star - \fo\|_{2, P_\phi(\nu) \times \pi_{\fh, Q^\star}}$, where $P_\phi(\nu) \times \pi_{\fh, Q^\star}$ is a distribution that also satisfies $\|(\cdot)/\mu\|_\infty \le C$ (Proposition~\ref{asm:con}) and hence can be handled by recursion. The third can be bounded by $\sqrt{C} \| \fo - \TG \fo\|_{2, \mu}$ due to $\|\nu/\mu\|_\infty \le C$, and $\|\fo - \TG \fo\|_{2, \mu}$ can be related to $\| \fo - \TG \fo\|_{2, \mu}$ by the concentration bounds established in Section~\ref{sec:concentration}, which are satisfied due to the choice of $|D|$ in the proposition statement. 

To prove Eq.\eqref{eq:hasgood}, we can similarly relate $\|\fo - \eTG \fo\|_{2, D}$ to $\|\fo - \TG \fo\|_{2, \mu}$ via the concentration bounds, and 
\begin{align*}
&~ \|\fo - \TG \fo\|_{2, \mu}
\le \|\fo - \TG \fo\|_{\infty} \\
\le &~ \|\fo - Q^\star\|_\infty + \|\TG Q^\star - \TG \fo\|_{\infty} \\
\le &~ (1+\gamma)\|\fo - Q^\star\|_\infty. \tag*{($\gamma$-contraction of $\TG$) \qedhere}	
\end{align*}
\end{proof}

\section{Proof of Theorem~\ref{thm:main}} \label{sec:tour}
With the careful analysis of the pairwise-comparison step given in Section~\ref{sec:subproblem}, we are now ready to analyze Algorithm~\ref{alg:bvft}. Roughly speaking, we will make the following arguments:
\begin{itemize}[leftmargin=*, topsep=0pt, itemsep=0pt]
\item For the output $\ff$, if $\max_{f'}\Ecal(\ff;f')$ is small, then $\ff \approx Q^\star$. (Eq.\eqref{eq:nobad} of Proposition~\ref{prop:phi})
\item That $\max_{f'}\Ecal(\ff;f')$ will be small, because $\max_{f'}\Ecal(f^\star; f')$ is small, where $f^\star\in\Fcal$ is the best approximation of $Q^\star$ in Definition~\ref{def:approx}. (Eq.\eqref{eq:hasgood} of Proposition~\ref{prop:phi})
\end{itemize}
Before we delve into the proof of Theorem~\ref{thm:main}, we need yet another lemma, which connects $\epsphi$ in Section~\ref{sec:subproblem} to the approximation error of $\Fcal$. As Section~\ref{sec:outline} has suggested, this is feasible because we are only concerned with the comparisons involving $f^\star$, and $\epsphi$ may be arbitrarily large otherwise. 

\begin{lemma} \label{lem:ffphi}
The $\phi$ induced from Line~7 satisfies $|\phi| \le (\Vmax/\epsdct)^2$. When $f^\star \in \{f, f'\}$, we further have $\epsphi \le \epsF + \epsdct$.
\end{lemma}

\begin{proof}[\textbf{Proof of Theorem~\ref{thm:main}}] 
Among the $(f,f')$ pairs enumerated in Lines~5 and 6, we will only be concerned with the cases when either $f=f^\star$ or $f'=f^\star$, and there are $2|\Fcal|$ such pairs. 
We require that w.p.~$1-\delta$,  Proposition~\ref{prop:phi} holds for all these $2|\Fcal|$ pairs. To guarantee so, we set the sample size $|D|$ to the expression in the statement of Proposition~\ref{prop:phi}, with $|\phi|$ replaced by its upper bound in Lemma~\ref{lem:ffphi} and $\delta$ replaced by $\delta/2|\Fcal|$ (union bound). We also let $\epsd = \epst$ to simplify the expressions, and will set the concrete value of $\epst$ later. The following is a sample size that satisfies all the above:
\begin{align} \label{eq:intermediate_sample_size}
|D| \ge \frac{82\Vmax^4 \ln\frac{160\Vmax|\Fcal|}{\epst \delta}}{\epst^2 \epsdct^2}.
\end{align}

Let $\phi$ be the partition induced by $\ff$ and $f^\star$.  According to Eq.\eqref{eq:nobad}, for any $\nu$ s.t.~$\|\nu/\mu\|_\infty \le C$,
\begin{align*}
\|\ff - Q^\star \|_{2,\nu} \le &~  \frac{2\epsphi + \sqrt{C}(\|\ff - \eTG \ff \|_{2, D} + 2\epst)}{1-\gamma} \\
= &~ \frac{2\epsphi + \sqrt{C}(\Ecal(\ff; f^\star) + 2\epst)}{1-\gamma} \\
\le &~ \frac{2\epsF + 2\epsdct + \sqrt{C}(\max_{f'\in\Fcal}\Ecal(\ff; f') + 2 \epst)}{1-\gamma}.
\end{align*}
It then remains to bound $\max_{f'}\Ecal(\ff; f')$. Note that
\begin{align*}
\max_{f' \in\Fcal}\Ecal(\ff; f') =   \min_{f \in \Fcal} \max_{f' \in \Fcal} \Ecal(f; f') 
\le  \max_{f' \in \Fcal} \Ecal(f^\star; f').
\end{align*}
For any $f'$, let $\phi'$ be the partition of $\Scal\times\Acal$ induced by $f^\star$ and  $f'$. Then
\begin{align*}
\Ecal(f^\star; f') = &~ \|f^\star - \eTGp f^\star\|_{2, D} \\
\le &~ (1+\gamma)\|f^\star - Q^\star\|_\infty + 2\epsphip + 2\epst   \tag{Eq.\eqref{eq:hasgood}} \\
\le &~ 4\epsF + 2\epsdct + 2 \epst.
\end{align*}
Combining the above results, we have for any $\nu$ s.t.~$\|\nu/\mu\|_\infty \le C$,
\begin{align*}
\|\ff - Q^\star \|_{2,\nu} 
\le &~ \frac{2\epsF + 2\epsdct + \sqrt{C}(4\epsF + 2\epsdct + 2\epst + 2\epst)}{1-\gamma} \\
\le &~ \frac{(2+4\sqrt{C})\epsF + 4\sqrt{C}(\epsdct + \epst)}{1-\gamma}.
\end{align*}
Finally, since any state-action distribution induced by any (potentially non-stationary) policy always satisfies Proposition~\ref{prop:phi}, by  \citet[Lemma 13]{chen2019information} we have
\begin{align*}
&~ J(\pi^\star) - J(\hat\pi) \le \frac{2}{1-\gamma} \sup_{\nu: \|\nu/\mu\|_\infty \le C} \|\ff - Q^\star\|_{\nu}  \\
\le &~ \frac{(4+8\sqrt{C})\epsF + 8\sqrt{C}(\epsdct + \epst)}{(1-\gamma)^2}.
\end{align*}
To guarantee that $\frac{8\sqrt{C}(\epsdct + \epst)}{(1-\gamma)^2} \le \epsilon \Vmax$, we set $\epsdct = \epst = \frac{(1-\gamma)^2 \epsilon \Vmax}{16\sqrt{C}}$. Plugging this back into Eq.\eqref{eq:intermediate_sample_size} yields the sample complexity in the theorem statement.
\end{proof}

\section{Discussions and Conclusions}


\subsection{Application to Model Selection} \label{sec:model_selection}
When learning $Q^\star$ from a batch dataset in practice, one would like to try different algorithms, different function approximators, and even different hyperparameters for a fixed algorithm and see which combination gives the best result, as is always the case in machine-learning practices. In supervised learning, this can be done by a simple cross-validation procedure on the holdout dataset. In batch RL, however, how to perform such a \emph{model-selection} step in a provably manner has been a widely open problem.\footnote{See \citet{mandel2014offline} and \citet{paine2020hyperparameter} for empirical advances on this problem.} 

There exists a limited amount of theoretical work on this topic, which often consider a restrictive setting when the base algorithms are model-based learners using nested state abstractions \citep{hallak2013model, seijen2014efficient,  jiang2015abstraction}.\footnote{An exception is the work of  \citet{farahmand2011model}, which requires the additional assumption that a regression procedure can approximate $\Tcal f$ and uses it to compute $\|f - \Tcal f\|$.} The only finite-sample guarantee we are aware of, given by \citet{jiang2015abstraction}, provides an oracle inequality with respect to an upper bound of $\|f - Q^\star\|$ based on how much the base state abstractions violate \emph{bisimulation} (or model-irrelevance) criterion \cite{whitt1978approximations, even-dar2003approximate, li2006towards} and $\ell_\infty$ concentration bounds, and the guarantee does not scale to the case where the number of base algorithms is super constant. 

In comparison, BVFT provides a more direct approach with a much stronger guarantee: let $Q_1, \ldots, Q_m$ be the output of different base algorithms. We can simply run BVFT on the holdout dataset with $\Fcal = \{Q_i\}_{i=1}^m$. The only function-approximation assumption we need is that one of $Q_i$'s is a good approximation of $Q^\star$, which is hardly an assumption as there is little we can do if all the base algorithms produce bad results. Compared to prior works, our approach is much more agnostic w.r.t.~the details of the base algorithms, our loss and guarantees are directly related to $\|f-Q^\star\|$ as opposed to relying on (possibly loose) upper bounds based on bisimulation, and our statistical guarantee scales to an exponentially large $\Fcal$ as opposed to a constant-sized one. 


Another common approach to model selection is to estimate $J(\pi)$ for each candidate $\pi$ via off-policy evaluation (OPE).\footnote{As a side note, BVFT can be adapted to OPE when $Q^\pi \in \Fcal$ for target policy $\pi$ as long as we change the $\max$ operator in $\eTG$ to $\pi$, though Assumption~\ref{asm:con} will still be needed.} OPE-based model selection has very different characteristics compared to BVFT, and they may be used together to complement each other; see a more detailed comparison and discussion in Appendix~\ref{app:ope}. 


\subsection{On the Assumption of Exploratory Data} \label{sec:fail}
As noted in Section~\ref{sec:data}, our Assumption~\ref{asm:con} adopts a relatively stringent definition of concentrability coefficient. A more standard definition is the following, as appeared in the hardness conjecture of \citet{chen2019information}:
\begin{assumption} \label{asm:con2}
Let $d^\pi_t$ be the distribution of $(s_t, a_t)$ when we start from $s_0 \sim d_0$ and follow policy $\pi$, which we will call an \emph{admissible distribution}. We assume that there exists $C < \infty$ such that $\|d_t^\pi/\mu\|_{\infty} \le C$ for any (possibly nonstationary) policy $\pi$ and $t\ge 0$. 
\end{assumption}
In Appendix~\ref{app:obstacle} we 
construct 3 scenarios to illustrate the difficulties (and sometimes possibilities) in extending our algorithm and its guarantees to a weaker data assumption such as Assumption~\ref{asm:con2}; due to space limit we only include a high-level summary of the results below. In the first construction, we show that BVFT fails under Assumption \ref{asm:con2} in a very simple MDP if we are allowed to provide a contrived $\mu$ distribution to the learner where data is unnaturally missing in certain states (Figure~\ref{fig:counterexample}). Motivated by the unnaturalness of the construction, we attempt to circumvent the hardness by imposing an additional mild assumption on top of Assumption~\ref{asm:con2}, that $\mu$ must itself be ``admissible'' . While it becomes much more difficult to construct a counterexample against the algorithm, it is still possible to design a scenario where our analysis breaks down seriously (Figure~\ref{fig:counter2}). We conclude with a positive result showing that the actual assumption we need is somewhere in between Assumptions~\ref{asm:con} and \ref{asm:con2}, for that our algorithm and analysis work for a simple and natural ``on-policy'' case which obviously violates Assumption~\ref{asm:con}; formulating a tighter version of the assumption in a natural and interpretable manner remains future work.

\subsection{Conclusions} 
We conclude the paper with a few open problems:
\begin{itemize}[leftmargin=*, topsep=0pt, itemsep=0pt]
\item Is it possible to circumvent the failure modes discussed in Section~\ref{sec:fail} with novel algorithmic ideas, so that a variant of BVFT only requires a weaker assumption on data? On a related note, the original hardness conjecture of \citet{chen2019information} remains unsolved: our positive result assumes a stronger data assumption, and the negative results of \citet{wang2020statistical, amortila2020variant} assume weaker ones.
\item 
When the data is seriously under-exploratory, to the extent that it is impossible to compete with $\pi^\star$ \citep{fujimoto2019off, liu2019off, liu2020provably}, what is the minimal function-approximation assumption that enables polynomial learning? In particular, requiring that $\Fcal$ realizes $Q^\star$ no longer makes sense as we do not even attempt to compete with $\pi^\star$. Recent works often suggest that we compete with $\pi$ whose occupancy is covered by $\mu$, but as of now very strong expressivity assumptions are needed to achieve such an ambitious goal \citep[e.g.,][Proposition 9]{jiang2020minimax}. It will be interesting to explore more humble objectives and see if the algorithmic and analytical ideas in this work extend to the more realistic setting of learning with non-exploratory data.
\end{itemize}


\section*{Acknowledgments}
The authors thank Akshay Krishnamurthy, Yu Bai, Yu-Xiang Wang for discussions related to Lemma~\ref{lem:distance_gen_err}, and Alekh Agarwal for numerous comments and discussions after the initial draft of the paper was released. Nan Jiang acknowledges support from the DEVCOM Army Research Laboratory under Cooperative Agreement W911NF-17-2-0196 (ARL IoBT CRA). The views and conclusions contained in this document are those of the authors and should not be interpreted as representing the official policies, either expressed or implied, of the Army Research Laboratory or the U.S.~Government. The U.S.~Government is authorized to reproduce and distribute reprints for Government purposes notwithstanding any copyright notation herein.

\bibliography{bib}
\bibliographystyle{icml2021}

\newpage
\appendix
\onecolumn
\section{Further Discussions of Assumption~\ref{asm:con}}
\subsection[Example of Bounded CS in Low-rank Environments]{Example of Bounded $\CS$ in Low-rank Environments} \label{app:CS}
Here we show that in general environments whose transition admits low-rank stochastic factorization, there always exists $\mu$ that satisfies Assumption~\ref{asm:con} with a small $C$. 

\begin{example}\label{exm:con}
Consider a low-rank MDP as defined in \citet[Proposition 9]{barreto2014policy, jiang2017contextual}, where the transition matrix $[P(s'|s,a)]_{(s,a), s'} = P_1 \times P_2$, and $P_1 \in \RR^{|\Scal\times\Acal|\times d}$ and $P_2 \in \RR^{d\times|\Scal|}$ are both row-stochastic matrices. Also assume that $d_0$ is a mixture of rows of $P_2$. Then, a data distribution $\mu$, where $\mu(s)$ is the average of $P_2$'s rows and $\mu(a|s)$ is uniform, satisfies  Assumption~\ref{asm:con} with $C \le d|\Acal|$. \footnote{We do not particularly attempt to satisfy $\mu(s,a) > 0~\forall s,a$ here, because it can always be satisfied by mixing in an infinitesimally small portion of $U(\Scal\times\Acal)$. In fact, this assumption itself can be removed from the main analysis with some additional care.}
\end{example}
\begin{proof}
$\CA\le |\Acal|$ follows from the uniformity of $\mu(a|s)$. For $\CS\le d$, note that $P(\cdot|s,a)$ and $d_0(\cdot)$ are convex combinations of rows of $P_2$, and $\mu(s)$ is designed to be uniform mixture of these rows, so $P(s'|s,a)/\mu(s')$ is always bounded by the number of rows, which is $d$. 
\end{proof}

\para{Comparison to Standard Concentrability} The more lenient and popular definitions of concentrability (e.g., Assumption~\ref{asm:con2}) are also found to be satisfiable in low-rank MDPs---in fact, such low-rankness is the only type of general structure known to enable concentrability \citep[Proposition 10]{chen2019information}. Comparing our Example~\ref{exm:con} with the example given by \citet{chen2019information}, the most outstanding difference is that in their case, the data distribution $\mu$ can be a mixture of state distributions induced by different policies in the environment; if a hidden factor cannot be reached by any policy, it is possible that any mixture distribution may fail to satisfy Assumption~\ref{asm:con} with a reasonably small $C$. 

\para{Comparison to No Inherent Bellman Errors} In the above low-rank MDP scenario, the assumption that $\Fcal$ has no \textit{inherent Bellman error} \cite{antos2008learning, chen2019information}---which enables polynomial sample complexity for many existing algorithms---can provably hold when the left factorization matrix (the analogy of $P_1$ in Example~\ref{exm:con}) is known to the learner as state-action features, so it is worth comparing such a setting to ours. In this setting, which is often known as \textit{linear MDPs} \cite{jin2020provably}, one can choose $\Fcal$ to be the linear class induced by the left factorization matrix, which is guaranteed to be closed under $\Tcal$, i.e., have no inherent Bellman errors.  In contrast, our Assumption~\ref{asm:con} holds without relying on knowing the left factorization matrix. 
The price we pay is that we require a stochastic factorization (Example~\ref{exm:con}) instead of just low-rankness, and whether Algorithm~\ref{asm:con} can hold with just low-rankness (possibly with additional mild assumptions) is an open problem. 


\subsection{Lower bounds} \label{app:lower}
\citet{wang2020statistical, amortila2020variant, zanette2020exponential, chen2021infinite} have recently proved hardness results under $Q^\star$ realizability in batch RL. Among them, the result of \cite{amortila2020variant} is most closely related to ours, and their setup is roughly that (1) $\Fcal$ is a linear function class (induced by feature map $\varphi$) with $Q^\star \in \Fcal$, and (2) the feature covariance matrix, $\EE_\mu[\varphi \varphi^\top]$, has lower-bounded eigenvalues under the data distribution. 
To understand how these results relate to ours, consider the following data assumption:
\begin{assumption}[$\Fcal$-aware Concentrability] \label{asm:conF}
We assume that there exists $C < \infty$, such that for any $f, f'\in\Fcal$, $\|f - f'\|_{2, d_t^\pi}^2 \le C\|f - f'\|_{2, \mu}^2 $ for any (possibly nonstationary) policy $\pi$ and $t\ge 0$ (see Assumption~\ref{asm:con2} for the definition of $d_t^\pi$). 
\end{assumption}
When $\Fcal$ is linear and $\EE_\mu[\varphi \varphi^\top]$ has large eigenvalues (assuming  $\|\phi\|_2\le 1$), Assumption~\ref{asm:conF} is automatically satisfied with $C$ being the inverse of the smallest eigenvalue of $\EE_{\mu}[\varphi \varphi^\top]$. Therefore, translating the lower bounds of \citet{wang2020statistical, amortila2020variant} to the setting of generic $\Fcal$, we have:
\begin{proposition}[Corollary of \citet{amortila2020variant}] \label{prop:lower}
In the setup of our main text, if we replace Assumption~\ref{asm:con} with Assumption~\ref{asm:conF}, no algorithm with finite sample complexity exists.
\end{proposition}
\begin{proof}
This is a direct corollary of \citet{amortila2020variant}, who show that finite sample complexity is unattainable even with stronger assumptions: linear $\Fcal$ is stronger than generic $\Fcal$, and  $\EE_{\mu}[\varphi \varphi^\top]$ having  eigenvalues lower-bounded away from $0$ is stronger than Assumption~\ref{asm:conF}. 
\end{proof}
Next we consider the relationship between Assumptions~\ref{asm:con}, \ref{asm:con2}, and \ref{asm:conF} in the following result; the proof is elementary and can be extracted from existing analyses \citep{munos2003error, munos2007performance, chen2019information}. 
\begin{proposition}
Assumption~\ref{asm:con} $\Rightarrow$ Assumption~\ref{asm:con2} $\Rightarrow$ Assumption~\ref{asm:conF}.
\end{proposition}
With these results, we now can relate the lower bounds \citet{wang2020statistical, amortila2020variant} to our positive result: indeed they do not contradict each other, as the negative results use the weakest form of concentrability (the $\Fcal$-aware version in Assumption~\ref{asm:conF}), and our positive result uses the strongest form (Assumption~\ref{asm:con}). Furthermore, to circumvent the hardness in Proposition~\ref{prop:lower}, imposing linear structure on $\Fcal$---which is a very strong structural assumption---does not help, as the hardness results of \citet{wang2020statistical, amortila2020variant} still apply. On the other hand, making a stronger data assumption as in Assumption~\ref{asm:con} would avoid the lower bound.

As a final remark, the hardness conjecture of \citet{chen2019information}, which uses Assumption~\ref{asm:con2}, remains unsolved. Originally \citet{chen2019information} argued that hardness conjecture is highly likely true given the lack of positive results under realizability, but given our work the picture is much less clear now. If we still anticipate a hardness result, our work has substantially narrowed the search space for the lower-bound constructions (if they exist): we will necessarily be able to establish the lower bound with either $|\Fcal|=2$ or $\Fcal$ being piecewise constant, otherwise our tournament procedure can extend the polynomial upper bounds for these special settings to arbitrary function classes. 

\section{Proofs} \label{app:proof}
\subsection{Proof of Lemma~\ref{lem:proj_bellman} }
Let $\Tcal_{M_\phi}$ denote the Bellman update operator of $M_\phi$. It suffices to show that  for any $f$, $\Tcal_{M_\phi} f \in \Fphi$ and is the 
argmin in Eq.\eqref{eq:TG}. $\Tcal_{M_\phi} f\in \Fphi$ follows directly as for any $s,a,\ts,\ta$ such that $\phi(s,a) = \phi(\ts,\ta)$, $R_\phi(s,a) = R_\phi(\ts,\ta)$ and $P_\phi(\cdot|s,a) = P_\phi(\cdot|\ts,\ta)$. For the other claim, note that projecting onto a piecewise constant function class means taking a weighted average within each partition, i.e., for any $s,a$,
\begin{align}
(\TG f)(s,a) = &~ \EE_{(\ts,\ta,r,s')\sim \mu}[r + \gamma V_f(s') | \phi(\ts,\ta) = \phi(s,a)] \\
= &~\frac{\sum_{\ts, \ta: \phi(\ts, \ta) = \phi(s,a)} \mu(\ts, \ta) (R(\ts,\ta)+ \gamma \EE_{s'\sim P(\ts,\ta)}[V_f(s')])}{\sum_{\ts, \ta: \phi(\ts, \ta) = \phi(s,a)} \mu(\ts, \ta) } \label{eq:Tphi} \\
= &~ R_\phi(s,a)+ \gamma \EE_{s'\sim P_\phi(s,a)}[V_f(s')] = (\Tcal_{M_\phi} f)(s,a). \tag*{\qed}
\end{align}

\subsection{Proof of Proposition~\ref{prop:epsphi0}}
The existence and the uniqueness of the fixed point of $\TG$ follow from Lemma~\ref{lem:proj_bellman}, so it suffices to check $Q^\star = \TG Q^\star$. For any $(s,a)$, we will calculate $(\TG Q^\star)(s,a)$ using Eq.\eqref{eq:Tphi}, which is a convex average of terms in the form of
\begin{align}
R(\ts,\ta)+ \gamma \EE_{s'\sim P(\ts,\ta)}[V_{Q^\star}(s')] = (\Tcal Q^\star)(\ts, \ta) = Q^\star(\ts,\ta),
\end{align}
where $\phi(\ts,\ta) = \phi(s,a)$. Since $\epsphi = 0$, we have $Q^\star(\ts, \ta) = Q^\star(s,a)$, and $(\TG Q^\star)(s,a)$ is the convex average of terms that are always equal to $Q^\star(s,a)$. Hence, $\TG Q^\star = Q^\star$. \qed

\subsection{Proof of Lemma~\ref{lem:Mphicon}}
$M$ and $M_\phi$ share the same initial state distribution $d_0$. Now for any $(s,a)$, $P_\phi(s'|s,a)$ is a convex combination of $P(s'|\ts,\ta)$ for $\{(\ts, \ta): \phi(s,a) = \phi(\ts, \ta)\}$, and since $\frac{P(s'|s,a)}{\mu(s')} \le \CS$ for every $(s,a)$, replacing the enumerator with any convex combination satisfies the same inequality. The condition about $\CA$ only concerns the data distribution $\mu$ and does not depend on the MDP. \qed

\subsection{Proof of Lemma~\ref{lem:qapprox}}
Let $g^\star = \argmin_{g\in\Fphi}\|g - Q^\star\|_\infty$, and $\|g^\star - Q^\star\|_\infty = \epsphi$. 
For any $(s,a)$, 
\begin{align*}
&~ \left|Q^\star(s,a) - (\TG Q^\star)(s,a)\right| \\
= &~ \left|Q^\star(s,a) -\frac{\sum_{\ts, \ta: \phi(\ts, \ta) = \phi(s,a)} \mu(\ts, \ta) (R(\ts, \ta) + \gamma \EE_{s' \sim P(\ts, \ta)} [V^\star(s')])}{\sum_{\ts, \ta: \phi(\ts, \ta) = \phi(s,a)} \mu(\ts, \ta) }\right|\\
= &~ \left|Q^\star(s,a) -\frac{\sum_{\ts, \ta: \phi(\ts, \ta) = \phi(s,a)} \mu(\ts, \ta) Q^\star(\ts, \ta)}{\sum_{\ts, \ta: \phi(\ts, \ta) = \phi(s,a)} \mu(\ts, \ta) }\right|
\le \max_{\ts, \ta: \phi(\ts, \ta) =  \phi(s,a)} |Q^\star(s,a) - Q^\star(\ts, \ta)|.
\end{align*}
Now for any $\ts, \ta$ s.t.~$\phi(\ts, \ta) = \phi(s,a)$, 
\begin{align*}
&~ |Q^\star(s,a) - Q^\star(\ts, \ta)| \\
= &~ |Q^\star(s,a) - g^\star(s,a) + g^\star(\ts, \ta) - Q^\star(\ts, \ta)| \tag{$g^\star\in\Fphi$ and is piecewise constant} \\
\le &~ |Q^\star(s,a) -g^\star(s,a)| + |g^\star(\ts, \ta) - Q^\star(\ts, \ta)| \le 2\epsphi. \tag*{\qed}
\end{align*}

\subsection{Proof of Lemma~\ref{lem:regression}} \label{app:least_square}
Fixing any $h\in\Hcal$, define random variable $Z(h) := (h(X) - Y)^2 - (h^\star(X) - Y)^2$, which has bounded range $[-\Vmax^2, \Vmax^2]$. Noting that $\EE[Z(h)] = \EE[(h(X) - h^\star(X))^2]$, we  bound the variance of $Z(h)$ as
\begin{align*}
\VV[Z(h)] \le &~ \EE[Z(h)^2] = \EE[\left((h(X) - Y)^2 - (h^\star(X) - Y)^2\right)^2] \\
= &~ \EE[(h(X) - h^\star(X))^2(h(X) + h^\star(X) - 2Y)^2]  \\
\le &~ 4\Vmax^2 \EE[(h(X) - h^\star(X))^2]
= 4\Vmax^2 \EE[Z(h)].
\end{align*}
Next, we derive a uniform derivation bound on $\EE[Z(h)] - \hat{\EE}[Z(h)]$ for all $h\in\Hcal$. Let $\Hcal' \subset \Hcal$ be the $\epsilon$-cover of $\Hcal$ under $\ell_\infty$. By the definition of covering number, we have that $|\Hcal'|=N$ and for any $h\in\Hcal$, there exists $h'\in\Hcal'$ such that $\|h - h'\|_\infty \le \epsilon$.  

Applying the one-sided Bernstein and union bounding over all $h'\in\Hcal'$: w.p.~$1-\delta$, $\forall h'\in H'$, 
\begin{align*}
\EE[Z(h')] - \hat{\EE}[Z(h')] 
\le &~ \sqrt{\frac{2 \VV[Z(h')] \log\frac{N}{\delta}}{n}} + \frac{4\Vmax^2 \log\frac{N}{\delta}}{3n} \\
\le &~ \sqrt{\frac{8\Vmax^2 \EE[Z(h)] \log\frac{N}{\delta}}{n}} + \frac{4\Vmax^2 \log\frac{N}{\delta}}{3n}.
\end{align*}
Now for any $h\in\Hcal$, let $h'$ be its closest function in $\Hcal'$, and 
\begin{align*}
\EE[Z(h)] - \hat{\EE}[Z(h)]  
= \EE[Z(h')] - \hat{\EE}[Z(h')] + (\EE[Z(h)] - \EE[Z(h')]) - (\hat\EE[Z(h)] - \hat\EE[Z(h')]). 
\end{align*}
We have already bounded the first term, so it suffices to bound the remaining two terms. Consider
\begin{align*}
|Z(h) - Z(h')|  = &~ |(h(X) - Y)^2 - (h'(X) - Y)^2| \\
= &~ |(h(X) - h'(X)) (h(X) + h'(X) - 2Y)| \le 2\Vmax \epsilon_0.
\end{align*}
So we have 
\begin{align*}
\EE[Z(h)] - \hat{\EE}[Z(h)]  
\le \sqrt{\frac{8\Vmax^2 \EE[Z(h)] \log\frac{N}{\delta}}{n}} + \frac{4\Vmax^2 \log\frac{N}{\delta}}{3n} + 4\Vmax \epsilon_0.
\end{align*}
Since $\hat{\EE}[(h(X)-Y)^2]$ and $\hat{\EE}[Z(h)]$ differ by a term that does not depend on $h$, the minimzer of the former, $\hat h$, also minimizes the latter. Therefore,
$$
\hat{\EE}[Z(\hat h)] \le \hat{\EE}[Z(h^\star)] = 0. 
$$
This leads to
\begin{align*}
\EE[(h(X)-h^\star(X))^2] = &~ \EE[Z(\hat h)]\le \sqrt{\frac{8\Vmax^2 \EE[Z(\hat h)] \log\frac{N}{\delta}}{n}} + \frac{4\Vmax^2 \log\frac{N}{\delta}}{3n} + 4\Vmax \epsilon_0.
\end{align*}
Solving for the quadratic formula, we have
\begin{align*}
&~ \EE[(h(X)-h^\star(X))^2] = \EE[Z(\hat h)] \\
\le &~ \left(\sqrt{\frac{2\Vmax^2 \log\frac{N}{\delta}}{n}} + \sqrt{\frac{10\Vmax^2 \log\frac{N}{\delta}}{3n} + 4\Vmax \epsilon_0}\right)^2 
\le  \frac{22\Vmax^2 \log\frac{N}{\delta}}{3n} + 8\Vmax\epsilon_0. \tag*{\qed}
\end{align*}

\subsection{Proof of Lemma~\ref{lem:generr}} \label{app:generr}
	For any $(s,a,r,s')$, let $x = \phi(s,a)$ and $y= r + \gamma V_f(s')$. When we sample $(s,a,r,s')$ according to the data distribution, we use $X$ and $Y$ to denote the random variables whose realizations are $x$ and $y$, respectively. Define $\Xcal$ as the codomain of $\phi$, and $|\Xcal|= |\phi|$. Consider the regression problem $x\mapsto y$ over function class $\Hcal = [0, \Vmax]^\Xcal$. Let $\hat h$ and $h^\star$ be the empirical risk minimizer and the Bayes-optimal regressor, respectively, and $h^\star \in \Hcal$ thanks to the full expressivity of $\Hcal$. Also note that $\Ncal_\infty(\Hcal, \epsilon_0) \le (\Vmax/\epsilon_0)^{|\Xcal|}$. Invoking Lemma~\ref{lem:regression} we immediately have that w.p.~$\ge 1-\delta/2$,
	\begin{align} \label{eq:udb}
	\EE_{\mu}[(\hat h(X) -  h^\star(X))^2] \le \frac{8\Vmax^2 \log\frac{2(\Vmax/\epsilon_0)^{|\phi|}}{\delta}}{|D|} + 8\Vmax\epsilon_0.
	\end{align}
	
	Next we establish $\EE_{\mu}[(\hat h(X) - h^\star(X))^2] = \|\eTG f - \TG f\|_{2, \mu}^2$. It suffices to show that for any $(s,a,r,s')$, $h^\star(\phi(s,a)) = (\TG f)(s,a)$ and $\hat h(\phi(s,a)) = (\eTG f)(s,a)$. We only show the former and the latter is similar. Let $x= \phi(s,a)$, and
	\begin{align*}
	&~ (\TG f)(s,a) = R_\phi(s,a) + \gamma \EE_{s'\sim P_\phi(s,a)}[V_f(s')] \\
	= &~ \frac{\sum_{\ts, \ta: \phi(\ts, \ta) = \phi(s,a)} \mu(\ts, \ta) (R(\ts, \ta) + \gamma \EE_{s' \sim P(\ts, \ta)}[V_f(s')])}{\sum_{\ts, \ta: \phi(\ts, \ta) = \phi(s,a)} \mu(\ts, \ta) } \\
	= &~ \EE_{(\ts,\ta)\sim\mu} [R(\ts, \ta) + \gamma \EE_{s' \sim P(\ts, \ta)}[V_f(s')] \,|\, \phi(\ts, \ta) = \phi(s,a)] \\
	= &~ \EE_{(\ts,\ta)\sim\mu, r = R(\ts, \ta), s' \sim P(\ts, \ta)} [r + \gamma V_f(s') \,|\, \phi(\ts, \ta) = x] \\
	= &~ \EE_{(s,a)\sim\mu, r = R(s, a), s' \sim P(s, a)} [Y \,|\, X = x]. \tag{change of variable} \\
	= &~ h^\star(x). 
	\end{align*}
	Finally we back up the sample size from the generalization error bound. To guarantee $\|\eTG f - \TG f\|_{2, \mu}^2 \le \epst^2$, we set $\epsilon_0 = \epst^2/16\Vmax$, and it suffices to have
	$$
	\frac{8\Vmax^2 \log\frac{(16\Vmax^2/\epst^2)^{|\phi|}}{\delta}}{|D|} \le \epst^2/2.
	$$
	Solving for $|D|$ completes the proof. \qed

\subsection{Proof of Lemma~\ref{lem:distance_gen_err}} \label{app:distance_gen_err}
We prove the concentration inequality in two separate cases depending on the magnitude of $\|\fo - g\|_{2, \mu}$. In the first case when $\|\fo - g\|_{2, \mu}$ is small, we may use Bernstein's inequality to obtain fast rate as $2\VV_{\mu}[( \fo(s, a) - g(s, a) )^2]$ is controlled by $\|\fo - g\|_{2, \mu}$. In the second case, large $\|\fo - g\|_{2, \mu}$ enables us to leverage the inequality of $|\|\fo - g\|_{2, D} - \|\fo - g\|_{2, \mu}| \leq |\|\fo - g\|_{2, D}^2 - \|\fo - g\|_{2, \mu}^2| / \|\fo - g\|_{2, \mu}$.
	
Before discussing these two cases separately, let's first establish some results as preparation. Let $\Fphi'$ be an $\epsilon_0'$-cover of $\Fphi$ under $\ell_\infty$, and $N \coloneqq \Ncal_{\infty}(\Fphi, \epsilon_0')$ be the covering number. Thus, for any $g\in \Fphi$, there exists $g' \in \Fphi'$ such that $\|g - g'\|_\infty \leq \epsilon_0'$. We also let $n = |D|$.
	
We apply Bernstein's inequality with a  union bound over $\Fphi'$: w.p.~$\ge 1-\delta/2$, for any $g' \in \Fphi'$,
\begin{align}
&~ \left|\|\fo - g'\|_{2, D}^2 - \|\fo - g'\|_{2, \mu}^2\right|
\\
= &~ \left|\frac{1}{n} \sum_{(s,a, r,s') \in D} \left(\fo(s, a) - g'(s, a)\right)^2 - \EE_{\mu}\left[ \left( \fo(s, a) - g'(s, a) \right)^2 \right] \right|
\\
\leq &~  \sqrt{\frac{2\VV_{\mu}\left[\left( \fo(s, a) - g'(s, a) \right)^2\right]\ln{\frac{4N}{\delta}}}{n}} + \frac{\Vmax^2 \ln{\frac{4N}{\delta}}}{3n} \tag{Bernstein's inequality}
\\
\label{eq:lem10sqerr_fst}
\leq &~  \sqrt{\frac{2 \Vmax^2 \|\fo - g'\|_{2, \mu}^2 \ln{\frac{4N}{\delta}}}{n}} + \frac{\Vmax^2 \ln{\frac{4N}{\delta}}}{3n},
\end{align}
where the lest inequality follows from the fact of $\VV_{\mu}[( \fo(s, a) - g'(s, a) )^2] \le \EE_{\mu}[(\fo-g')^4] \leq \|(\fo - g')^2\|_\infty \EE_\mu[(f_0 - g')^2]$.

\paragraph{Case 1: $\|\fo - g\|_{2, \mu}$ is small}

Now, for any $g \in \Fphi$, let $g' \in \Fphi'$ satisfies $\|g - g'\|_\infty \leq \epsilon_0'$. 
\begin{align}
&~ \left|\|\fo - g\|_{2, D} - \|\fo - g\|_{2, \mu}\right|
\\
\leq &~ \left|\|\fo - g'\|_{2, D} - \|\fo - g'\|_{2, \mu}\right| + 2\epsilon_0'
\\
\leq &~ \sqrt{\left|\|\fo - g'\|_{2, D}^2 - \|\fo - g'\|_{2, \mu}^2 \right|} + 2\epsilon_0' \tag{$|a - b|^2 \leq |a^2 - b^2|$ for $a,b \geq 0$}
\\
\leq &~ \sqrt{\sqrt{\frac{2 \Vmax^2 \|\fo - g'\|_{2, \mu}^2 \ln{\frac{4N}{\delta}}}{n}} + \frac{\Vmax^2 \ln{\frac{4N}{\delta}}}{3n}} + 2\epsilon_0' \tag{Eq.\eqref{eq:lem10sqerr_fst}}
\\
\leq &~ \sqrt[4]{\frac{2 \Vmax^2 \|\fo - g'\|_{2, \mu}^2 \ln{\frac{4N}{\delta}}}{n}} + \sqrt{\frac{\Vmax^2 \ln{\frac{4N}{\delta}}}{3n}} + 2\epsilon_0' \tag{$\sqrt{a + b} \leq \sqrt{a} + \sqrt{b}$ for $a,b \geq 0$}
\\
\label{eq:lem10fneq_fst}
\leq &~ \sqrt[4]{\frac{2 \Vmax^2 \|\fo - g\|_{2, \mu}^2 \ln{\frac{4N}{\delta}}}{n}} + \sqrt[4]{\frac{2 \Vmax^2 \epsilon_0'^2 \ln{\frac{4N}{\delta}}}{n}} +  \sqrt{\frac{\Vmax^2 \ln{\frac{4N}{\delta}}}{3n}} + 2\epsilon_0'.
\end{align}

\paragraph{Case 2: $\|\fo - g\|_{2, \mu}$ is large}

Similarly, for any $g \in \Fphi$, let $g' \in \Fphi'$ satisfies $\|g - g'\|_\infty \leq \epsilon_0'$. Then, as long as $\|\fo - g\|_{2, \mu} \ne 0$,
\begin{align}
&~ \left|\|\fo - g\|_{2, D} - \|\fo - g\|_{2, \mu}\right| 
\leq \frac{\left|\|\fo - g\|_{2, D}^2 - \|\fo - g\|_{2, \mu}^2\right|}{\|\fo - g\|_{2, \mu}}
\\
\leq &~ \underbrace{\frac{\left|\|\fo - g'\|_{2, D}^2 - \|\fo - g'\|_{2, \mu}^2\right|}{\|\fo - g\|_{2, \mu}}}_{\text{(I)}} + \underbrace{\frac{4 \|\fo - g\|_{2, \mu} \epsilon_0' + 2 \left|\|\fo - g\|_{2, D} - \|\fo - g\|_{2, \mu}\right| \epsilon_0' + 2 \epsilon_0'^2}{\|\fo - g\|_{2, \mu}}}_{\text{(II)}}.
\end{align}
The last line is obtained by the following argument: 
\begin{align}
&~ \left| \left|\|\fo - g\|_{2, D}^2 - \|\fo - g\|_{2, \mu}^2\right| - \left|\|\fo - g'\|_{2, D}^2 - \|\fo - g'\|_{2, \mu}^2\right| \right|
\\
\leq &~ \left|\left(\|\fo - g\|_{2, D}^2 - \|\fo - g'\|_{2, D}^2\right) - \left(\|\fo - g\|_{2, \mu}^2 - \|\fo - g'\|_{2, \mu}^2 \right)\right|
\\
\leq &~ \left|\|\fo - g\|_{2, D}^2 - \|\fo - g'\|_{2, D}^2 \right| + \left| \|\fo - g\|_{2, \mu}^2  - \|\fo - g'\|_{2, \mu}^2\right|
\\
= &~ \left|\|\fo - g\|_{2, D} - \|\fo - g'\|_{2, D} \right|\cdot \left|\|\fo - g\|_{2, D} + \|\fo - g'\|_{2, D} \right| \\
&~ + \left| \|\fo - g\|_{2, \mu}  - \|\fo - g'\|_{2, \mu}\right| \cdot \left| \|\fo - g\|_{2, \mu}  + \|\fo - g'\|_{2, \mu}\right|
\\
\leq &~ \epsilon_0' \cdot (2\|\fo - g\|_{2, D} + \epsilon_0') + \epsilon_0' \cdot (2\|\fo - g\|_{2, \mu} + \epsilon_0')
\\
\leq &~ 4\|\fo - g\|_{2, \mu} \epsilon_0' + 2\left| \|\fo - g\|_{2, D} - \|\fo - g\|_{2, \mu} \right| \epsilon_0' + 2 \epsilon_0'^2,
\end{align}
where all the inequalities follow from the triangle inequality and the fact of $\|g - g'\|_\infty \leq \epsilon_0'$.

We now analyze the two terms above separately.
\begin{align}
\text{(I)} \leq &~ \frac{1}{\|\fo - g\|_{2, \mu}} \left(\sqrt{\frac{2 \Vmax^2 \|\fo - g'\|_{2, \mu}^2 \ln{\frac{4N}{\delta}}}{n}} + \frac{\Vmax^2 \ln{\frac{4N}{\delta}}}{3n} \right) \tag{by Eq.\eqref{eq:lem10sqerr_fst}}
\\
\leq &~ \frac{1}{\|\fo - g\|_{2, \mu}} \left(\left(\|\fo - g\|_{2, \mu} + \epsilon_0'\right)\sqrt{\frac{2 \Vmax^2 \ln{\frac{4N}{\delta}}}{n}} + \frac{\Vmax^2 \ln{\frac{4N}{\delta}}}{3n} \right)
\\
\label{eq:lem10term1}
\leq &~ \sqrt{\frac{2 \Vmax^2 \ln{\frac{4N}{\delta}}}{n}} + \frac{1}{\|\fo - g\|_{2, \mu}} \left( \sqrt{\frac{2 \epsilon_0'^2 \Vmax^2 \ln{\frac{4N}{\delta}}}{n}} + \frac{\Vmax^2 \ln{\frac{4N}{\delta}}}{3 n} \right).
\end{align}

\begin{align}
\label{eq:lem10term2}
\text{(II)} = &~ 4 \epsilon_0' + \frac{2 \epsilon_0'}{\|\fo - g\|_{2, \mu}} \left|\|\fo - g\|_{2, D} - \|\fo - g\|_{2, \mu}\right| + \frac{2 \epsilon_0'^2}{\|\fo - g\|_{2, \mu}}.
\end{align}

\paragraph{Combine them all}

We now unify those two cases above. We first set $\epsilon_0' = \epsd / 20$, and $N = \Ncal_{\infty}(\Fphi, \epsilon_0') \leq (\Vmax/\epsilon_0')^{|\phi|}$. 

When $\|\fo - g\|_{2, \mu} < 4 \epsilon_0'$, we apply the first case and obtain
\begin{align}
&~ \left|\|\fo - g\|_{2, D} - \|\fo - g\|_{2, \mu}\right|
\\
\leq &~ \sqrt[4]{\frac{2 \Vmax^2 \|\fo - g\|_{2, \mu}^2 \ln{\frac{4N}{\delta}}}{n}} + \sqrt[4]{\frac{2 \Vmax^2 \epsilon_0'^2 \ln{\frac{4N}{\delta}}}{n}} +  \sqrt{\frac{\Vmax^2 \ln{\frac{4N}{\delta}}}{3n}} + 2\epsilon_0'
\\
\leq &~ \left( \sqrt[4]{\frac{2}{25}} + \sqrt[4]{\frac{1}{200}} \right) \sqrt[4]{\frac{\Vmax^2 \epsd^2 \ln{\frac{4N}{\delta}}}{n}} + \sqrt{\frac{\Vmax^2 \ln{\frac{4N}{\delta}}}{3n}} + \frac{\epsd}{10}
\\
\leq &~ \left( \sqrt[4]{\frac{2}{25}} + \sqrt[4]{\frac{1}{200}} \right) \sqrt[4]{\frac{\Vmax^2 \epsd^2 |\phi| \ln{\frac{80 \Vmax}{\epsd \delta}}}{n}} + \sqrt{\frac{\Vmax^2 |\phi| \ln{\frac{80 \Vmax}{\epsd \delta}}}{3n}} + \frac{\epsd}{10}.
\end{align}

Solving
\begin{align}
\left( \sqrt[4]{\frac{2}{25}} + \sqrt[4]{\frac{1}{200}} \right) \sqrt[4]{\frac{\Vmax^2 \epsd^2 |\phi| \ln{\frac{80 \Vmax}{\epsd \delta}}}{n}} + \sqrt{\frac{\Vmax^2 |\phi| \ln{\frac{80 \Vmax}{\epsd \delta}}}{3n}} + \frac{\epsd}{10} \leq \epsd
\end{align}
implies that it suffices to set
\begin{align}
|D| = n \geq \frac{16 \Vmax^2 |\phi| \ln{\frac{80 \Vmax}{\epsd \delta}}}{\epsd^2},
\end{align}
for the case of $\|\fo - g\|_{2, \mu} < 4 \epsilon_0' = \frac{1}{5} \epsd$.

If $\|\fo - g\|_{2, \mu} \geq 4 \epsilon_0'$, we use the second case. The term (I) in Eq.\eqref{eq:lem10term1} is
\begin{align}
\text{(I)} \leq \frac{5}{4} \sqrt{\frac{2 \Vmax^2 \ln{\frac{4N}{\delta}}}{n}} + \frac{\Vmax^2 \ln{\frac{4N}{\delta}}}{12 \epsilon_0' n},
\end{align}
and the term (II) in Eq.\eqref{eq:lem10term2} is
\begin{align}
\text{(II)} \leq \frac{9}{2} \epsilon_0' + \frac{1}{2} \left|\|\fo - g\|_{2, D} - \|\fo - g\|_{2, \mu}\right|.
\end{align}

Since $\left|\|\fo - g\|_{2, D} - \|\fo - g\|_{2, \mu}\right| \leq \text{(I)} + \text{(II)}$, we reorder the terms and obtain
\begin{align}
\left|\|\fo - g\|_{2, D} - \|\fo - g\|_{2, \mu}\right|
\leq   \frac{5}{2} \sqrt{\frac{2 \Vmax^2 \ln{\frac{4N}{\delta}}}{n}} + \frac{\Vmax^2 \ln{\frac{4N}{\delta}}}{6 \epsilon_0 n} + 9\epsilon_0.
\end{align}
We still set $\epsilon_0' = \epsd / 20$. Thus, solving
\begin{align}
\frac{5}{2} \sqrt{\frac{2 \Vmax^2 \ln{\frac{4N}{\delta}}}{n}} + \frac{\Vmax^2 \ln{\frac{4N}{\delta}}}{6 \epsilon_0' n} + 9\epsilon_0' = \frac{5}{2} \sqrt{\frac{2 \Vmax^2 |\phi| \ln{\frac{80 \Vmax}{\epsd \delta}}}{n}} + \frac{10 \Vmax^2 |\phi| \ln{\frac{80 \Vmax}{\epsd \delta}}}{3 \epsd n} + \frac{9 \epsd}{20}\leq \epsd
\end{align}
provides us with the sufficient sample size $|D|$,
\begin{align}
|D| = n \geq \frac{50 \Vmax^2 |\phi| \ln{\frac{80 \Vmax}{\epsd \delta}}}{\epsd^2}.
\end{align}
Choosing the greater one between the required sample sizes for $\|\fo - g\|_{2, \mu} < 4 \epsilon_0' = \epsd / 5$ and $\|\fo - g\|_{2, \mu} \geq 4 \epsilon_0' = \epsd / 5$ completes the proof. \qed

\subsection{Proof of Proposition~\ref{prop:phi}}
Since $|D|$ in the proposition statement is chosen to satisfy the sample-size requirements in Lemmas~\ref{lem:generr} and \ref{lem:distance_gen_err}, the statement of each lemma holds with probability at least $1-\delta/2$, and by union bound they hold simultaneously w.p.~$\ge 1-\delta$. 
	
\paragraph{Bounding $\|Q^\star - \fo\|$ using $\|\fo - \eTG \fo\|_{2, D}$ (Eq.\eqref{eq:nobad}):} ~\\
Define $\pi_{f, f'}$ as the policy $s \mapsto \argmax_{a} \max\{f(s,a), f'(s,a)\}$. Consider any $\nu$ such that $\|\nu/\mu\|_\infty \le C$,
\begin{align}
\|Q^\star - \fo\|_{2, \nu} \le &~ \|Q^\star - \TG Q^\star\|_{2, \nu} + \|\TG Q^\star - \TG \fo\|_{2, \nu} +  \|\fo- \TG \fo\|_{2, \nu}  \\
\le &~ 2\epsphi + \gamma \|Q^\star - \fo\|_{2, P_\phi(\nu) \times \pi_{\fh, Q^\star}}+ \sqrt{C} \| \fo - \TG \fo\|_{2, \mu}. \label{eq:error_prop}
\end{align}
In Eq.\eqref{eq:error_prop}, the first term follows from Lemma~\ref{lem:qapprox} and that $\|\cdot\|_{2, \nu} \le \|\cdot\|_\infty$, the second from \citet[Lemmas 14 and 15]{chen2019information}, and the third from \citet[Lemma 12]{chen2019information}.
	
According to Lemma~\ref{lem:Mphicon}, $P_\phi(\nu) \times \pi_{\fh, Q^\star}$ also satisfies $\|(\cdot)/\mu\|_{\infty} \le C$, so it can be viewed as one of those $\nu$'s we started with on the LHS, allowing us to expand the inequality indefinitely. Alternatively, we have
\begin{align*}
&~ \sup_{\nu: \|\nu/\mu\|_\infty \le C} \|Q^\star - \fo\|_{2, \nu} \\
\le &~ \gamma \sup_{\nu: \|\nu/\mu\|_\infty \le C} \left(\|Q^\star - \fo\|_{2, P_\phi(\nu) \times \pi_{\fo, Q^\star}} \right)+ 2\epsphi +\sqrt{C} \| \fo - \TG \fo\|_{2, \mu} \\
\le &~ \gamma \sup_{\nu: \|\nu/\mu\|_\infty \le C} \left(\|Q^\star - \fo\|_{2, \nu} \right)+ 2\epsphi +\sqrt{C} \| \fo - \TG \fo\|_{2, \mu} \tag{Lemma~\ref{lem:Mphicon}}.
\end{align*}
So for any $\nu$ such that $\|\nu/\mu\|_\infty \le C$, $\|Q^\star - \fh \|_{2, \nu} \le \frac{2\epsphi + \sqrt{C}\| \fo - \TG \fo\|_{2, \mu}}{1-\gamma}$. 
	
It then remains to bound $\| \fo - \TG \fo\|_{2, \mu}$:
\begin{align*}
\| \fo - \TG \fo\|_{2, \mu} 
\le &~ \| \fo - \eTG \fo\|_{2, \mu} + \|\eTG \fo - \TG \fo\|_{2, \mu}  \\
\le &~ \| \fo - \eTG \fo\|_{2, \mu} + \epst \tag{Lemma~\ref{lem:generr}}\\
\le &~ \| \fo - \eTG \fo\|_{2, D} + \epsd + \epst. \tag{Lemma~\ref{lem:distance_gen_err}}
\end{align*}
	
\paragraph{Bounding $\|\fo - \eTG \fo\|_{2, D}$ using $\|Q^\star -\fo\|$ (Eq.\eqref{eq:hasgood}):}
\begin{align*}
\|\fo - \eTG \fo\|_{2, D} \le &~ \|\fo - \eTG \fo\|_{2, \mu} + \epsd \tag{Lemma~\ref{lem:distance_gen_err}}\\
\le &~ \|\fo - \TG \fo\|_{2, \mu} + \epst + \epsd \tag{Lemma~\ref{lem:generr}}\\
\le &~ \|\fo - \TG \fo\|_{\infty} + \epst + \epsd \tag{$\|\cdot\|_{2, \mu} \le \|\cdot\|_{\infty}$}\\
\le &~ \|\fo - Q^\star\|_\infty + \|\TG Q^\star - \TG \fo\|_{\infty} + \|Q^\star - \TG Q^\star\|_\infty + \epst + \epsd \\
\le &~ (1+\gamma)\|\fo - Q^\star\|_\infty + 2\epsphi + \epst + \epsd. \tag*{($\gamma$-contraction of $\TG$ under $\ell_\infty$ (Lemma~\ref{lem:proj_bellman}) and Lemma~\ref{lem:qapprox}) \qed}
\end{align*}

\subsection{Proof of Lemma~\ref{lem:ffphi}}
For the first claim, we may write $\phi(s,a) = (\bar{f}(s,a), \bar{f'}(s,a))$, and the number of equivalent classes induced by $\phi$ is at most the product of the cardinalities of the codomains of $\bar{f}$ and $\bar{f'}$, so the result follows.
	
For the second claim, recall that $\epsphi = \epsilon_{\Fphi} = \min_{g\in\Fphi} \|g - Q^\star\|_\infty$, so 
\begin{align}
\epsphi \le &~ \|\bar{f^\star} - Q^\star\|_\infty \tag{$f^\star \in \{f, f'\}$ and $\bar{f^\star} \in \Fphi$} \\
\le &~ \|f^\star - Q^\star\|_\infty + \|\bar{f^\star} - f^\star\|_\infty 
\le \epsF + \epsdct. \tag*{\qed}
\end{align}

\section{Obstacles in Relaxing Assumption~\ref{asm:con}} \label{app:obstacle}
In this section we discuss the obstacles in relaxing Assumption~\ref{asm:con}. Before we start, we emphasize that the difficulties have nothing to do with the tournament procedure, \textbf{and are entirely about learning $Q^\star$ with a realizable state-action aggregation $\phi$}---a problem so standard, that the difficulties we find may have broader implications beyond the scope of this work; see Appendix~\ref{app:counter} for discussions on the relevance of our findings to existing RL algorithms.

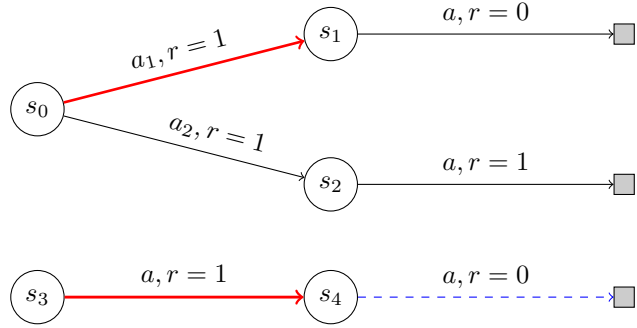
\begin{figure}
	\centering

\begin{tikzpicture}
\node[circle,
minimum width =20pt ,
minimum height =20pt ,draw=black] (0) at(0,2){$s_0$};
\node[circle,
minimum width =20pt ,
minimum height =20pt ,draw=black] (5) at(0,-0.5){$s_3$};
\node[circle,
minimum width =20pt ,
minimum height =20pt ,draw=black] (6) at(3.9,-0.5){$s_4$};
\node[rectangle,
minimum width =7.63932pt ,
minimum height =7.63932pt ,draw=black, fill=black!20] (7) at(7.8,-0.5){};
\node[circle,
minimum width =20pt ,
minimum height =20pt ,draw=black] (2) at(3.9,1){$s_2$};
\node[circle,
minimum width =20pt ,
minimum height =20pt ,draw=black] (1) at(3.9,3){$s_1$};
\node[rectangle,
minimum width =7.63932pt ,
minimum height =7.63932pt ,draw=black, fill=black!20] (8) at(7.8,1){};
\node[rectangle,
minimum width =7.63932pt ,
minimum height =7.63932pt ,draw=black, fill=black!20] (9) at(7.8,3){};

\draw[color=red,line width=1pt, ->] (0) --node[color=black,above,sloped]{$a_1,r = 1$}(1);
\draw[->] (0) --node[color=black,above,sloped,pos=0.618]{$a_2,r = 1$}(2);
\draw[->] (1) --node[color=black,above,sloped]{$a,r = 0$}(9);
\draw[color=red, line width=1.08pt, ->] (5) --node[color=black,above,sloped]{$a,r = 1$}(6);
\draw[->] (2) --node[color=black,above,sloped]{$a,r = 1$}(8);
\draw[dashed,color=blue,->] (6) --node[color=black,above,sloped]{$a,r = 0$}(7);

\end{tikzpicture}
	\caption{MDP construction for Section~\ref{sec:fail}; see Appendix~\ref{app:counter} for details. 
		$s_0$ is the deterministic initial state. The gray squares represent known absorbing states.  Actions $a_1, a_2$ have the same effects (and abbreviated as ``$a$'') in all states except in $s_0$. The {\bf \color{red} red thick} arrows for $(s_0, a_1)$ and $(s_3, a)$ indicate that these state-action pairs are aggregated together (they share the same $Q^\star$ value). The {\color{blue}  blue dashed} arrow for $(s_4, a)$ indicates that data is missing for this state-action pair, which violates Assumption~\ref{asm:con}; Assumption~\ref{asm:con2} still holds since no policy can visit $s_3$ or $s_4$ from $s_0$. The missing data leads to high uncertainty in $Q^\star(s_3, a)$, and such uncertainty propagates to $Q^\star(s_0, a_1)$ and causes the failure of learning. \label{fig:counterexample}}
\end{figure}

\paragraph{Negative Result: Contrived $\mu$}
We present our first counterexample in Figure~\ref{fig:counterexample}, where  Assumption~\ref{asm:con} is violated but
Assumption~\ref{asm:con2} is satisfied due to missing data in state-action pairs unreachable from the initial state $s_0$. 
A state-action aggregation $\phi$, which is guaranteed to express $Q^\star$, results in a projected Bellman operator $\TG$ that has multiple fixed points other than $Q^\star$ even when sample size goes to infinity, and many such fixed points produce suboptimal policies. Therefore, $\|f - \TG f\|_{2, \mu}$, which is the surrogate loss that plays a central role in our algorithm, cannot control the performance of $\pi_f$ in this setting; see Appendix~\ref{app:counter} for details. 

\paragraph{Negative Result: Admissible $\mu$}
An issue with Figure~\ref{fig:counterexample} is that its $\mu$ cannot be admitted in the MDP (i.e., generated by some behavior policy from $d_0$), and assuming admissible $\mu$ (which is reasonable) excludes this counterexample. While this may look promising, here we show that our analysis still faces substantial obstacles if we replace Assumption~\ref{asm:con} with Assumption~\ref{asm:con2} plus admissible $\mu$. Below we explain in more details. 

As Section~\ref{sec:err_prop} has alluded to, the error propagates according to the dynamics of $M_\phi$ instead of $M$ in our analysis, so the purpose of Assumption~\ref{asm:con} is really to guarantee the following type of assumption:\footnote{The actual assumption we need is slightly more complicated, that we need $\|\nu/\mu\|_{\infty} \le C$ for any $\nu$ admissible in $M_\phi$ but using any other admissible $\nu'$ from $M$ as the initial distribution. This complication, however, does not affect our counterexample. Neither does it affect the next positive result.}
\begin{assumption} \label{asm:actual}
	There exists $C < \infty$, such that for any $\phi$, we have $\|\nu/\mu\|_{\infty} \le C$ for any $\nu$ that is admissible in $M_\phi$.
\end{assumption} 
Unfortunately, via a carefully constructed example, we show in Figure~\ref{fig:counter2} that Assumption~\ref{asm:actual} cannot be implied by Assumption~\ref{asm:con2} plus admissible $\mu$. In particular, even when Assumption~\ref{asm:con2} is satisfied with a constant $C$ and $\mu$ is admissible, we can use the aggregation to ``leak'' probabilities from easy-to-reach states in $\mu$ to hard-to-reach states gradually over time steps, causing an exponential blow-up of $\|\nu/\mu\|_\infty$. 
See Appendix~\ref{app:counter2} for details.

\tikzstyle{Default}=[fill=white, draw=black, shape=circle]
\tikzstyle{Rec}=[fill=white, draw=black, shape=rectangle]

\tikzstyle{Unidirectional}=[->]
\tikzstyle{Double Arrow}=[<->]
\tikzstyle{Line}=[-]
\tikzstyle{Dashed arrow}=[->, dashed]
\tikzstyle{Dashed double arrow}=[<->, dashed]
\tikzstyle{dashed line}=[-, dashed]
\tikzstyle{blue line}=[-, color=blue]
\tikzstyle{Dotted Arrow}=[->, dotted]

\begin{figure}[t]
	\centering
	\begin{tikzpicture}
	\begin{pgfonlayer}{nodelayer}
		\node [style=Default] (0) at (-6.5, 5) {};
		\node [style=Default] (1) at (-6.5, 5) {$s_1$};
		\node [style=Default] (2) at (-7.5, 3.25) {};
		\node [style=Default] (3) at (-5.5, 3.25) {$s_2'$};
		\node [style=none] (12) at (-6, 3.75) {};
		\node [style=none] (13) at (-6, 2.75) {};
		\node [style=none] (14) at (-3, 2.75) {};
		\node [style=none] (15) at (-3, 3.75) {};
		\node [style=Default] (17) at (-3.75, 3.25) {$s_2$};
		\node [style=Default] (18) at (-4.75, 1.5) {};
		\node [style=Default] (19) at (-2.75, 1.5) {$s_3'$};
		\node [style=Default] (25) at (-1.25, 1.5) {$s_3$};
		\node [style=Default] (26) at (-2.25, 0) {};
		\node [style=Default] (27) at (-0.25, 0) {$s_4'$};
		\node [style=none] (29) at (-2.5, 5.75) {$d_0$};
		\node [style=none] (31) at (0.5, 0.75) {$\cdots$};
		\node [style=none] (32) at (-4.75, 5.5) {$1-p$};
		\node [style=none] (33) at (-4, 4.75) {$p(1-p)$};
		\node [style=none] (34) at (-0.75, 4.75) {$p^2(1-p)$};
		\node [style=none] (35) at (-3.5, 2) {};
		\node [style=none] (36) at (-3.5, 1) {};
		\node [style=none] (37) at (-0.5, 1) {};
		\node [style=none] (38) at (-0.5, 2) {};
	\end{pgfonlayer}
	\begin{pgfonlayer}{edgelayer}
		\draw [style=Unidirectional] (1) to (2);
		\draw [style=Unidirectional] (1) to (3);
		\draw [style=blue line] (13.center) to (14.center);
		\draw [style=blue line] (14.center) to (15.center);
		\draw [style=blue line] (15.center) to (12.center);
		\draw [style=blue line] (12.center) to (13.center);
		\draw [style=Unidirectional] (17) to (18);
		\draw [style=Unidirectional] (17) to (19);
		\draw [style=Unidirectional] (25) to (26);
		\draw [style=Unidirectional] (25) to (27);
		\draw [style=Dotted Arrow, bend right=15, looseness=1.25] (29.center) to (17);
		\draw [style=Dotted Arrow, in=0, out=-180, looseness=1.50] (29.center) to (1);
		\draw [style=Dotted Arrow, bend left] (29.center) to (25);
		\draw [style=blue line] (36.center) to (37.center);
		\draw [style=blue line] (37.center) to (38.center);
		\draw [style=blue line] (38.center) to (35.center);
		\draw [style=blue line] (35.center) to (36.center);
	\end{pgfonlayer}
\end{tikzpicture}
	\caption{The example in Figure~\ref{fig:counterexample} requires that $\mu$ visits unreachable states. While assuming that $\mu$ can be naturally generated in the MDP using a behavior policy (``admissible'') would exclude the pathology in Figure~\ref{fig:counterexample}, here we show that in a more carefully constructed example, dynamics of $M_\phi$ allow us to visit a state exponentially more likely than in $\mu$, while Assumption~\ref{asm:con2} is satisfied with a constant $C$. Roughly speaking, $\mu(s_t)\propto p^t$ but it is possible to produce a distribution in $M_\phi$ that visits $s_t$ w.p.$\propto (2p)^t$ for $p\ll 1$; see Appendix~\ref{app:counter2}.  \label{fig:counter2}}
\end{figure}
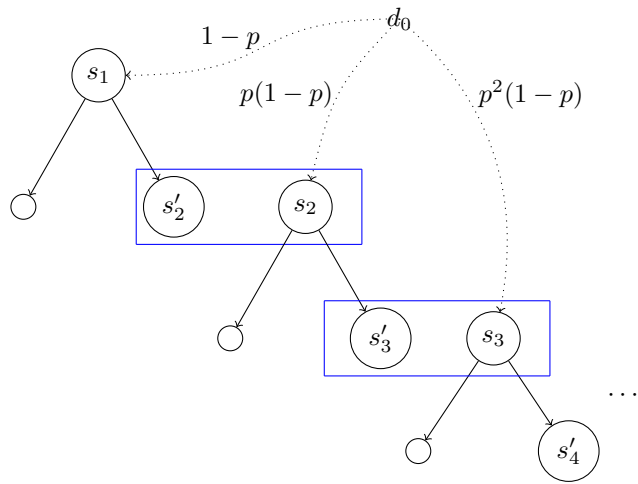

\paragraph{Positive Result: On-policy Data}
Despite the discouraging counterexamples, we show a slightly positive result, implying that Assumption~\ref{asm:actual} \emph{could} be much weaker than Assumption~\ref{asm:con}, leaving the possibility of something weaker than Assumption~\ref{asm:con} but more natural and interpretable than Assumption~\ref{asm:actual}. 
In particular, we show a scenario where Assumptions~\ref{asm:con2} and \ref{asm:actual} can be satisfied with a small $C$, yet Assumption~\ref{asm:con} may be violated badly, implying the looseness and unnecessity of Assumption~\ref{asm:con} for our analysis.

Consider the uncontrolled case where there is only one action, and we may treat $P$ as an $|\Scal|\times|\Scal|$ transition matrix. We consider the ``on-policy'' case, where $d_0=\mu$ is an invariant distribution w.r.t.~$P$, i.e., $\mu^\top P = \mu^\top$. Since an invariant distribution always exists, this does not impose any restriction on $P$, so we can make the $C$ in Assumption~\ref{asm:con} very large: for example, when $P$ is identity, $C$ must be as large as $|\Scal|$ to satisfy Assumption~\ref{asm:con}. On the other hand, Assumption~\ref{asm:con2} is trivially satisfied with $C=1$, as $\mu$ is the only admissible distribution in $M$. Perhaps surprisingly, this is also true for Assumption~\ref{asm:actual}: regardless of $\phi$, we have $\mu^\top P_\phi = \mu^\top P = \mu$, because $P_\phi$ is defined by averaging the dynamics of $P$ with weights proportional to $\mu$, and this averaging step can be ignored when the incoming distribution is $\mu$ itself.

\subsection{Details of Figure~\ref{fig:counterexample}}
\label{app:counter}
We construct an MDP, a data distribution $\mu$, and a realizable function class $\Fcal$ with $|\Fcal|=2$, such that
(1) Assumption~\ref{asm:con} is violated; 
(2) Assumption~\ref{asm:con2} is satisfied; 
(3) Algorithm~\ref{alg:bvft} may output a suboptimal policy even with infinite data.
	
See Figure~\ref{fig:counterexample} for an illustration of the MDP, where the transition dynamics and the rewards are deterministic. Let $\gamma = 0.9$, and $s_0$ be the deterministic initial state. 
	
Let $\mu$ be uniform over all state-action pairs other than $(s_4,a)$,\footnote{The probability assigned to $(s_3, a)$ is twice as much as that to $(s_0, a_1)$ by $\mu$, because the former is the abbreviation of two state-action pairs. This detail is of minor importance, and $\mu$ can be changed to many other distributions, as long as the later values of $Q$ are set in a way consistent with $\mu$.} which violates Assumption~\ref{asm:con}.\footnote{The fact that $\mu(s,a)>0~ \forall s,a$ is violated is of minor concern here: we can add exponentially small probabilities to $(s_4, a)$ in $\mu$, and with a  polynomially large $D$, the non-uniqueness of the fixed point of $\eTG$ still persists. What is really important is that no finite $\CS$ satisfies $P(s'|s,a)/\mu(s') \le \CS$.} However, since no policy can visit $s_3$ or $s_4$ from the starting state $s_0$, lacking data in $s_4$ does not affect the validity of Assumption~\ref{asm:con2}. It remains to specify $\Fcal$ and show that our algorithm fails.
	
\newcommand{\rr}[1]{{\color{red}#1}}
\newcommand{\bb}[1]{{\color{blue}#1}}

Our $\Fcal$ consists of two functions, $Q^\star$ and $Q$. We specify them by writing down their values on $\rr{(s_0, a_1)}, (s_0, a_2), (s_1, a), (s_2, a), \rr{(s_3, a)}, \bb{(s_4, a)}$ as a vector: $Q^\star = (\rr{1},1.9,0,1,\rr{1},\bb{0})$ and $Q = (\rr{7}, 1.9, 0, 1, \rr{7}, \bb{10})$. The red and blue colors correspond to the color schemes in Figure~\ref{fig:counterexample} to facilitate understanding.
	
When we run Algorithm~\ref{alg:bvft}, the only nontrivial aggregation $\phi$ performs is grouping together $(s_0, a_1)$ and $(s_3, a)$; $(s_0, a_2)$ and every other state-action pair are kept in their own equivalence classes, respectively. ($a_1, a_2$ under the same state are by default aggregated except in $s_0$.)
	
Now that the construction is complete, we can verify that our loss $\|f - \TG f\|$ is zero for both $f=Q^\star$ and $f = Q$. 
Therefore, the algorithm may choose to output the greedy policy of either function, but the greedy policy of $Q$ is suboptimal since it chooses $a_1$ in $s_0$. 

\paragraph{Broader Implications} The non-uniqueness of the fixed point of $\TG$ not only affects our algorithm, but also affects many popular algorithms when they are applied with the piecewise constant class induced by $\phi$, as \emph{any} fixed point of $\TG$ is a valid output of these algorithms. Such algorithms include iterative algorithms such as FQI and (both $\ell_1$ and $\ell_2$) Bellman residual minimization-style algorithms \citep{xie2020q}. Moreover, related algorithms that do not directly approximate $Q^\star$ but instead perform policy iteration \citep[e.g., LSPI;][]{lagoudakis2003least} are also subject to this counterexample, as the $Q^\pi$ learned at the policy evaluation step is equal to $Q^\star$ regardless of the policy in this MDP.\footnote{This is because the only action choice is in the initial state $s_0$.} 

This phenomenon is particularly interesting when we notice that, everything will work fine if we remove the data on $(s_3, a)$: the algorithm will still have high uncertainty in the values of $s_3$ and $s_4$, but such uncertainty does not incorrectly propagate to $s_0$ and hence does not affect our ability to choose the optimal action there. Therefore, the pathological behavior is due to having data with \emph{more} coverage than necessary, which may be surprising. To our best knowledge, this pathology---which affects a wide range of batch RL algorithms---is documented for the first time. It will be interesting to see if we can obtain a deeper understanding of this issue and possibly circumvent it.

\subsection{Counterexample Against Admissible $\mu$} \label{app:counter2}
One weakness of the counterexample in Figure~\ref{fig:counterexample} is that $\mu$ cannot be generated by a behavior policy starting from $d_0$, as such \emph{admissible distributions}  never visit $s_3$. Therefore, we can exclude the pathology by assuming that $\mu$ is (a mixture) of admissible distributions, on top of Assumption~\ref{asm:con2}. 
\begin{assumption}[$\mu$ is admissible] \label{asm:con3}
In addition to Assumption~\ref{asm:con2}, assume that $\mu$ is a mixture of $d_t^\pi$ for a set of $(\pi, t)$ pairs, where $\pi$ may be nonstationary and/or stochastic. 
\end{assumption}
This seemingly mild additional assumption leads to some powerful corollaries. For example, any distribution induced from $\mu$ (say, with policy $\pi'$ in $t'$ steps) as the initial distribution (instead of $d_0$) will still be covered by $\mu$, since the distribution is essentially a mixture of $d_{\pi\circ \pi'}^{t+t'}$. Furthermore, creating a situation like Figure~\ref{fig:counterexample} becomes very difficult (see below for detailed reasons); in fact, we believe that it is impossible to induce hardness with a constant-depth construction as in Figure~\ref{fig:counterexample}. 

Despite the power of Assumption~\ref{asm:con3}, we show below that our analysis still faces substantial obstacles even under Assumption~\ref{asm:con3}. In particular, our proof around Eq.\eqref{eq:error_prop} requires that distribution induced in $M_\phi$ should also be well covered by $\mu$ (i.e., Assumption~\ref{asm:actual}), and this is guaranteed by Assumption~\ref{asm:con} via Lemma~\ref{lem:Mphicon}. If we replace Assumption~\ref{asm:con} with Assumption~\ref{asm:con3}, however, below we show in a counterexample that $M_\phi$ can induce a distribution that visits a state exponentially more likely compared to its likelihood in $\mu$, thus breaking Assumption~\ref{asm:actual}.

\paragraph{Counterexample} See Figure~\ref{fig:counter2}. The true MDP $M$ is essentially a 2-armed contextual bandit, where the initial state (or context) distribution is $d_0(s_t) = (1-p)p^{t-1}$ for $t$ between $1$ and a sufficiently large integer $N$ (and the rest probability goes to $s_{N+1}$; we will not need it). $0<p<1$ is a parameter to be set later. For each $s_t$, we will call the two actions $L$ (for ``left'') and $R$ (``right''), respectively. All states other than $\{s_t\}$ are absorbing states. During data collection, the learner randomly starts in some $s_t$ according to $d_0$, and take actions uniformly at random. This guarantees that $C=2$ for Assumption~\ref{asm:con2}. Note that we do not specify the rewards as our goal is only to show that a distribution with large density ratio against $\mu$ can be induced in $M_\phi$, and not to directly show the failure of the algorithm.

\paragraph{Aggregation and Dynamics of $M_\phi$} The dynamics of $M_\phi$ (Definition~\ref{def:Mphi}) can be equivalently described as the following: given $(s,a)$, the next-state $s'$ is generated in two steps,
\begin{enumerate}
\item Re-draw $(\ts, \ta)$ from $\{(\ts, \ta): \phi(\ts, \ta) = \phi(s,a)\}$ with probability proportional to $\mu$. 
\item Sample $s'\sim P(\ts, \ta)$. 
\end{enumerate}
In Figure~\ref{fig:counter2}, we aggregate $s_t'$ and $s_t$ together for every $t$, and the ``re-drawing'' step happens after the agent lands in $s_t'$. That is, before the next time step, there is some probability that it will teleport to $s_t$. Strictly speaking we are aggregating states instead of state-action pairs, but the effects can be reproduced by e.g., adding a dummy state with only 1 action above each $s_t$, and aggregating this state-action pair with $(s_{t-1}, R)$. We opt for a slightly different mechanism for simplicity of the construction.

\paragraph{Calculation} We will show that with the effect of aggregation, we can induce a distribution whose density ratio against $\mu$ is exponentially large, using the policy that always takes action ``$R$''. Let $P(s_t)$ be the probability of visiting $s_t$ at time step $t-1$ by this policy, and let $\mu(s_t)$ be the mass of $s_t$ in $\mu$. Our goal is to calculate $P(s_t)/\mu(s_t)$ and show that it is exponential in $t$. 
First, $\mu(s_t) = p^{t-1}(1-p)/2$. The division by $2$ is because $\mu$ contains data from two different time steps ($\{s_t\}$ at time step $0$, and $\{s_{t'}\}$ and their sibling states at time step $1$). Note that the data is collected in the true MDP $M$ and the aggregation plays no role here.

Now we calculate $P(s_t)$. The only probability path of visiting $s_t$ at time step $t-1$ is $s_1 \to s_2' \to s_2 \to s_3' \rightarrow s_3 \to \ldots \to s_t' \rightarrow s_t$. Along this path, $P(s_1) = 1-p$, and $s_{t-1} \to s_t'$ is deterministic, so we focus on the probability of $s_t' \to s_t$. 

Recall from the above that whenever at $s_t'$, we will redraw a state from $\{s_t', s_t\}$ according to their probabilities in $\mu$. Therefore,
$$
P(s_t' \to s_t) = \frac{\mu(s_t)}{\mu(s_t') + \mu(s_t)} = \frac{p^{t-1}(1-p) / 2}{p^{t-2}(1-p) / 4 + p^{t-1}(1-p) / 2} = \frac{2p}{1+2p},
$$
where $\mu(s_t')$ is calculated based on the fact that the data collection policy is uniformly random. This gives
$$
P(s_t) = (1-p) \left(\frac{2p}{1+2p}\right)^{t-1}, \quad \frac{P(s_t)}{\mu(s_t)} = \frac{1}{2}\left(\frac{2}{1+2p}\right)^{t-1}.
$$
Therefore, when $p < 1/2$, $P(s_t)/\mu(s_t)$ will be exponential in $t$.

\section{What If Assumption~\ref{asm:con} is Violated?} \label{app:violate}
When Assumption~\ref{asm:con} is violated, the guarantees in Theorem~\ref{thm:main} does not hold in general. However, this does not mean that the algorithm is useless or there is nothing we can do in this case. Below we discuss a few actionable items and briefly sketch the theoretical analyses that justifies them.

\subsection[Diagnosing the Uniqueness of T's Fixed Point]{Diagnosing the Uniqueness of $\eTG$'s Fixed Point}
As Section~\ref{sec:fail} has shown, one possible consequence of not having Assumption~\ref{asm:con} is that the fixed point of $\eTG$ may be non-unique. This suggests a diagnostic procedure that checks if such pathology occurs: let $\widehat{\Fphi} := \{g \in\Fphi: \|g - \eTG g\| \le \epsilon'\}$ be the set of approximate fixed points of $\eTG$ where $\epsilon'$ is some small threshold. Then, we may compute 
$\max_{g,g'\in\widehat{\Fphi}} \|g - g'\|_{2, D}$ as a statistic for the diagnosis. If Assumption~\ref{asm:con} is satisfied, such a maximum distance should be small as all functions in $\widehat{\Fphi}$ should be close to the fixed point of $\TG$ under $\|\cdot\|_{2, \mu}$, \footnote{In an early stage of the project, our algorithm actually checks whether $f \in \widehat{\Fphi}$ instead of the current Line~8. This alternative algorithm also enjoys polynomial sample complexity.} and it is important to note that such a claim does not depend on $f^\star \in \{f, f'\}$ (see e.g.,~Lemma~\ref{lem:proj_bellman}). On the other hand, if the maximum distance within $\widehat{\Fphi}$ is observed to be small when we actually run the algorithm, we can rest assured that $\|\ff - Q^\star\|_{2, \mu}$ is small (assuming $\max_{f' \in \Fcal} \Ecal(\ff;f')$ is small), and we only need Assumption~\ref{asm:con2} to further guarantee the near-optimality of $\pi_{\ff}$, regardless of whether Assumption~\ref{asm:con} holds or not. 

As an example, consider the counterexample in Figure~\ref{fig:counterexample} and Appendix~\ref{app:counter}, where both $Q^\star$ and $Q$ are fixed points of $\TG$ and $\|Q^\star - Q\|_{2, \mu}$ is large. As Section~\ref{sec:fail} suggests, the pathology goes away if we remove the data from $(s_3, a)$. Note that $\TG$ still has many fixed points (the value of $(s_4, a)$ can be anything), but their distance from each other under $\|\cdot\|_{2, \mu}$ is always $0$ because $\mu$ is only supported on $s_0, s_1, s_2$, and all these fixed points induce an optimal policy from $s_0$.

\subsection[Tweaking phi]{Tweaking $\phi$}
Our main analysis treats the discretization step (Line~7) very casually. However, the structure of $\phi$ plays an important role in error propagation, and small changes in discretization can produce significantly different $\phi$'s, so one may want to search for a favorable $\phi$ among all possibilities. What should be the guideline for such a search?

We propose searching for $\phi$ that maximizes the least number of data points that fall into a group of state-action pairs.\footnote{One may recall that this is also a preferred situation in the analyses of state abstractions \citep{paduraru2008model, jiang2015abstraction, nan_abstraction_notes}, as $\ell_\infty$ quantities are only statistically stable when each abstract state receives enough data. However, this is not the primary reason for our recommendation: our statistic $\|f - \eTG f\|_{2, D}$ is stable and generalizes well regardless of the structure of $\phi$, thanks to the use of weighted $\ell_2$ norm instead of $\ell_\infty$ norm; see Section~\ref{sec:concentration}.}
If each group of state-action pairs receives enough data, $\mu_{\min}^\phi  := \sum_{\ts, \ta: \phi(\ts, \ta) = \phi(s,a)} \mu(\ts, \ta)$ will be lower bounded away from $0$ for any $(s,a)$, which guarantees well-controlled error propagation. To see this, recall that Eq.\eqref{eq:error_prop} is the key step that characterizes distribution shift error propagation, and we need Assumption~\ref{asm:con} and Lemma~\ref{lem:Mphicon} to upper bound $\|\fo - \TG \fo\|_{2, \nu}$ with $\sqrt{C} \|\fo - \TG\fo\|_{2, \mu}$ for any $\nu$ produced by the dynamics of $M_\phi$. If $\mu_{\min}^\phi$ is large, however, we do not need to reason about the properties of $\nu$. Instead, the fact that both $\fo$ and $\TG \fo$ are (approximately) in $\Fphi$ (recall that $\phi$ is produced by $f$ and $f'$ in Line~7 with $\fo=f$) and $\mu_{\min}^\phi$ is lower bounded already allows us to state the desired bound with $C$ replaced by $1/\mu_{\min}^\phi$, up to small additive errors. Therefore, if we can always guarantee a lower-bounded $\mu_{\min}^\phi$ (via the sample size in each group as a surrogate) whenever in Line~7, the algorithm enjoys its sample complexity guarantee \emph{without any concentrability assumptions}, which is very appealing. 

Of course, there is no guarantee that we can find such a well-behaved $\phi$ for a single pair of $f,f'$, let alone the $|\Fcal|^2$ pairs that all need to be handled. Therefore, this suggestion is more suitable for the model-selection scenario where $|\Fcal|$ is small (Section~\ref{sec:model_selection}). To produce a rich set of possible $\phi$'s, one can consider various designs of the discretization grid (see Footnote~\ref{ft:discrete}): changing the offset of the grid, using a non-regular grid, using soft aggregations instead of hard ones, or even setting $\epsdct$ to be slightly greater than intended. Whether it is easy to find a well-behaved $\phi$ is more of an empirical question, and we leave further investigation to future work.

\section[Defining epsilonF in Weighted L2 Norm]{Defining $\epsF$ in Weighted $\ell_2$ Norm} \label{app:l2approx}
It is possible to define $\epsF$ as $\inf_{f \in \Fcal}\|f-Q^\star\|_{2, \mu}$, and still prove a polynomial sample complexity result. However, making this change leads to a suboptimal dependence on $C$ if we still follow the same proof structure. Below we briefly explain the challenges.

In Section~\ref{sec:subproblem} we characterize $\phi$ with only two parameters, $|\phi|$ and $\epsphi$, and $\epsphi$ is later bounded in Lemma~\ref{lem:ffphi} by $\epsF + \epsdct$. When we define $\epsF$ as $\inf_{f \in \Fcal}\|f-Q^\star\|_{2, \mu}$, the $\epsF$ part of $\epsphi$ is subject to the $\sqrt{C}$ penalty when there is distribution shift, but the $\epsdct$ part of $\epsphi$ is an $\ell_\infty$ discretization error and intuitively should not be affected by distribution shift. However, since we bundle $\epsF$ and $\epsdct$ together in $\epsphi$, $\epsdct$ will suffer an additional  $\sqrt{C}$ penalty compared to our current analysis, leading to additional dependence on $C$ in the sample complexity. 

One possible solution is to define $\epsF$ as $\inf_{f \in \Fcal}\sup_{\nu: \|\nu/\mu\|_\infty \le C}\|f-Q^\star\|_{2, \nu}$, i.e., to capture distribution shift inside the definition of $\epsF$, so that $\epsF$ is still effectively (a soft version of) $\ell_\infty$ error. The same change needs to be made in the definition of $\epsphi$. 
The difficulty with this approach is that such a ``soft $\ell_\infty$'' $\epsF$ does not play well with Lemma~\ref{lem:qapprox}: to avoid $\epsphi$ being amplified by $\sqrt{C}$, we need to also replace the LHS of Lemma~\ref{lem:qapprox} with $\sup_{\nu} \|Q^\star - \TG Q^\star\|_{2, \nu}$. However, it is unclear whether this quantity can be bounded by $\epsphi$ without paying $\sqrt{C}$, as  $\TG Q^\star =\argmin_{g \in\Fphi} \|g - \Tcal Q^\star\|_{2, \mu}= \argmin_{g \in\Fphi} \|g - Q^\star\|_{2, \mu}$ is the best approximation of $Q^\star$ in $\Fphi$ under weighting $\mu$, but ideally we want the best approximation under worst-case $\nu$ to avoid $\sqrt{C}$ penalty. We leave the resolution of this technical issue to future work. 

\section{Comparison to OPE in Model Selection}
\label{app:ope}
Section~\ref{sec:model_selection} discussed the application of BVFT to model selection. 
Another common approach to model selection is to estimate $J(\pi)$ for each candidate $\pi$ via off-policy evaluation (OPE). Unfortunately, unbiased OPE with importance sampling (IS) \citep{precup2000eligibility} incurs exponential variance in horizon when the behavior policy is significantly different from the ones being evaluated \citep{jiang2016doubly}, and recent marginalized IS (MIS) methods that overcome such a ``curse of horizon'' require some nontrivial function-approximation assumptions. For example, even if one has a function class $\Qcal$ that realizes $Q^\pi$ for \emph{every} $\pi$ being evaluated, the state-or-the-art approaches only provide an interval that contains $J(\pi)$ without tightness guarantees \citep{jiang2020minimax, feng2020accountable}, and tightness requires further assumptions on realizing marginalized importance weights \citep[e.g.,][]{liu2018breaking, uehara2020minimax}. On a related note, if a rich $\Qcal$ is used to better satisfy these additional assumptions, one has to reserve a large amount of data as holdout dataset due to the statistical complexity of $\Qcal$. In comparison, BVFT only uses function classes of a well-controlled worst-case complexity $O(1/\epsilon^2)$.

Despite the additional assumptions, OPE has its own advantages: OPE directly estimates $J(\pi)$ instead of going through $\|f - Q^\star\|$ as a surrogate, and as a consequence, it removes the assumption that the base algorithms need to approximate $Q^\star$ and hence enjoys wider applicability. The information about $J(\pi)$ can also be valuable in certain application scenarios, which cannot be obtained by our approach (we can at the best provide an upper bound on $J(\pi^\star) - J(\pi)$). To this end, OPE-based methods and BVFT have very different characteristics when applied to model selection, and they are complementary and may be used together to provide more information and help the practitioners in making the final decision. 


\end{document}